\documentclass[10pt]{extarticle}

\usepackage[a4paper,margin=1in]{geometry}

\usepackage[T1]{fontenc}
\usepackage[scaled=0.95]{helvet}

\usepackage{microtype}

\usepackage{setspace}
\usepackage{amsmath,amssymb}
\usepackage{booktabs}
\usepackage{multirow}
\usepackage{array}
\usepackage{graphicx}
\usepackage{adjustbox}
\usepackage{float}
\usepackage[labelfont=bf,font=small,skip=6pt]{caption}
\DeclareCaptionLabelSeparator{nature}{ \textbar{} }
\usepackage[italic,defaultmathsizes]{mathastext}

\usepackage{titlesec}
\titleformat{\section}{\normalfont\Large\bfseries}{\thesection}{0.6em}{}
\titleformat{\subsection}{\normalfont\large\bfseries}{\thesubsection}{0.6em}{}
\titleformat{\subsubsection}{\normalfont\normalsize\bfseries\itshape}{\thesubsubsection}{0.6em}{}
\titleformat{\paragraph}[runin]{\normalfont\normalsize\bfseries}{}{0pt}{}[.\hspace{0.5em}]
\titlespacing*{\section}{0pt}{1.6em}{0.5em}
\titlespacing*{\subsection}{0pt}{1.1em}{0.35em}
\titlespacing*{\subsubsection}{0pt}{0.8em}{0.3em}
\titlespacing*{\paragraph}{0pt}{0.6em}{0.5em}
\usepackage[super,comma,sort&compress]{natbib}
\usepackage[colorlinks=true,linkcolor=black,citecolor=black,urlcolor=black]{hyperref}

\begin{document}

\noindent{\LARGE\bfseries Learning Normal Representations for Blood Biomarkers\par}
\vspace{1.0em}

\noindent Aashna P. Shah$^{1,2,6}$, Michelle M. Li$^{1,6}$, Yash Lal$^{4}$, Seffi Cohen$^{1,6,7}$, Liat F. Antwarg$^{1,6,7}$, Morgan Sanchez$^{1,6}$, James A. Diao$^{1,3,6}$, Chirag J. Patel$^{1}$, Ben Y. Reis$^{1,5,6,7}$, Ran D. Balicer$^{6,7}$*, Noa Dagan$^{6,7,8}$*, and \\ Arjun K. Manrai$^{1,6}$*

\noindent{\footnotesize
\begin{list}{\arabic{enumi}.}{%
  \usecounter{enumi}\setlength{\leftmargin}{1.6em}\setlength{\labelsep}{0.5em}%
  \setlength{\itemindent}{0pt}\setlength{\listparindent}{0pt}%
  \setlength{\itemsep}{1pt}\setlength{\parsep}{0pt}\setlength{\topsep}{3pt}}
\item Department of Biomedical Informatics, Harvard Medical School, Boston, MA, USA
\item Department of Systems Biology, Harvard Medical School, Boston, MA, USA
\item Department of Medicine, Brigham and Women's Hospital, Boston, MA, USA
\item Department of Mathematics, Johns Hopkins University, Baltimore, MD, USA
\item Computational Health Informatics Program (CHIP), Boston Children's Hospital, Boston, MA, USA
\item The Ivan and Francesca Berkowitz Family Living Laboratory Collaboration at Harvard Medical School and Clalit Research Institute, USA and Israel
\item Clalit Research Institute, Innovation Division, Clalit Health Services, Ramat-Gan, Israel
\item Faculty of Computer and Information Science, Ben Gurion University, Be'er Sheva, Israel
\end{list}
}

\noindent{\small * Co-senior authors\\
Correspondence: \href{mailto:Arjun_Manrai@hms.harvard.edu}{Arjun\_Manrai@hms.harvard.edu}}

\section{Abstract}\label{abstract}

Blood-based biomarkers underpin clinical diagnosis and management, yet their interpretation relies largely on fixed population reference intervals that ignore stable, intra-patient variability. As such, population-based interpretation can mask meaningful deviation from an individual's baseline, risking delayed disease detection. To remedy this, there have been increasing efforts to personalize blood biomarker interpretation using individual testing histories. However, these methods may overfit to sparse data, inflating false-positive rates and unnecessary follow-up, and can also unwittingly include unrecognized or subclinical disease. Here, we leverage nearly 2 billion longitudinal laboratory measurements from over 1.6 million individuals across North America, the Middle East, and East Asia, to show that while laboratory values are highly individual, purely personalized intervals routinely overfit, classifying up to 68\% of measurements as abnormal, without corresponding associations with adverse clinical outcomes. We then introduce NORMA, a conditional transformer-based framework that generates reference intervals by conditioning on both a patient's history and population-level data about ``normal'' variation. NORMA-derived intervals achieve higher precision for predicting outcomes, including mortality, acute kidney injury, and chronic disease. These findings caution against over-personalization in laboratory medicine and demonstrate that anchoring individual trajectories to population-level priors outperforms either approach alone. To promote transparency, we publicly release the model, code, and an interactive user interface for accessible, individualized laboratory interpretation.

\section{Introduction}\label{main}

Laboratory testing is among the most frequently performed medical tests, with more than 14 billion tests ordered annually in the United States alone\cite{Freedman2015-wh}. Yet interpretation has changed remarkably little since the 1960s; values are classified as ``low,'' ``normal,'' or ``high'' against fixed population reference intervals (Pop$_{RI}$) that represent the central 95\% of measurements from ostensibly healthy individuals\cite{Guyatt1992-yd,Newsome2017-ya,Inzucchi2012-ci,Sikaris2017-in,Ceriotti2009-lm,Friedberg2007-zq,Muskens2022-nk}. Prior studies have recognized that many routine biomarkers fluctuate around narrow individual ``setpoints'' and that within-person change carries prognostic information that population intervals miss\cite{Ference2017-gw,Buergel2022-kz,Giannini2005-gx,Walters1996-ax,Boyd2010-cz,Manrai2018-mv,Nazir1999-sm,Foy2025-dj,Wang2022-qg,Coskun2022-pp,Coskun2021-xa,Obstfeld2021-ew}. Despite this, clinical practice often still relies on universal thresholds, partly because of the ease of a one-size-fits-all approach, and because operationalizing individualized interpretation requires modeling patient-specific trajectories at scale.

Efforts to move beyond universal thresholds have taken two directions. The first refines reference intervals for predefined subgroups, for example, sex-specific hematologic ranges or adjusted HbA1c thresholds for patients with hemoglobin variants\cite{Lacy2017-ct,Feingold2025-md,Walters1996-ax,Addo2021-ak,Rushton2001-tx,Manrai2018-mv}. Defining population subgroups introduces several challenges---such groups are often defined coarsely or arbitrarily, may encode existing biases, and cannot capture the full spectrum of individual variation embedded in longitudinal trajectories. The practice of race adjustment, in particular, has been increasingly challenged with major clinical and non-clinical implications for patients\cite{Diao2024-lw, Vyas2020-hf, Shah2025-ev}. The second approach derives personalized reference intervals (Per$_{RI}$) directly from each patient's laboratory testing history. Foy et al., for example, showed that deviations from individually derived hematologic setpoints associate more strongly with mortality than deviations from population thresholds, underscoring the prognostic value of within-person change\cite{Foy2025-dj}. However, purely individualized approaches that derive baselines entirely from a patient's own data risk incorporating unrecognized chronic disease into the estimated setpoint and overfitting to sparse histories, producing overly narrow intervals that label benign physiological variation as pathological or vice versa\cite{Foy2025-dj,Coskun2022-pp,Tayob2022-vz,Kohane2006-tq}.

Here, we ask whether these competing approaches can be combined using nearly 2 billion measurements across 30 analytes, spanning two decades of national longitudinal care, dense ICU monitoring, and perioperative data across three countries (Fig.~\ref{fig:figure1}). We first quantify the individuality of routine blood tests and then systematically compare population and personalized reference intervals. We then introduce NORMA (Normal Outcome Range Modeling with Attention), a new conditional transformer framework that combines both approaches to generate individualized reference intervals by anchoring each patient's trajectory to population-level expectations for a healthy state. Rather than choosing between the false dichotomy of purely population and personalized approaches, NORMA learns where each patient's interval is expected to fall between the two extremes. We show that purely personalized intervals overcall abnormalities and lose prognostic signal, that NORMA resolves this tradeoff, and test whether the resulting intervals detect clinically meaningful change months to years before population intervals.

\section{Results}
\subsection{Study populations}\label{study-populations}

We assessed population-based (Pop$_{RI}$), personalized (Per$_{RI}$), and NORMA-derived (NORMA$_{RI}$) reference intervals across 30 routine laboratory analytes spanning four clinical panels: complete blood count, comprehensive metabolic panel, lipid panel, and hemoglobin A1c (HbA1c). These panels are used in routine health evaluation, cardiometabolic risk assessment, and diabetes screening or monitoring, as detailed in the Methods. Within this analyte set, NORMA was trained on 3.4 million longitudinal laboratory sequences from two development cohorts: MIMIC-IV (179,601 patients) and EHRSHOT (5,676 patients); no development data were used in external validation (Supplementary Tables~\ref{tab:analyte_reference}--\ref{tab:norma_design}).

We validated NORMA in three independent cohorts that differed in temporal scale and clinical setting. The Clalit Health Services (CHS) cohort included 1,450,862 adults (63\% female; mean age 55.1 ± 18.6 years) contributing approximately 1.9 billion laboratory measurements over more than two decades of outpatient follow-up (Fig.~\ref{fig:figure2}a; Supplementary Table~\ref{tab:norma_design}). To ensure stable baseline estimation, we required at least five outpatient measurements per analyte spaced at least 90 days apart, yielding trajectories spanning a median of more than four years.

The eICU Collaborative Research Database (eICU-CRD) cohort included 98,432 ICU patients (46\% female; mean age 63.3 ± 16.1 years) across 208 U.S. hospitals, contributing over 20 million measurements with dense sampling (median number of measurements per patient ranged between 8-35 for high-frequency analytes) over a median stay of approximately 8 days.

The INSPIRE cohort included 51,159 surgical patients (52\% female; mean age 57.6 ± 14.9 years) undergoing perioperative care in South Korea, contributing over 10 million laboratory measurements collected across preoperative, intraoperative, and postoperative timepoints, with a median follow-up of approximately 175 days.

Laboratory distributions in CHS, EHRSHOT, and INSPIRE largely overlapped population reference intervals, whereas eICU-CRD and MIMIC-IV exhibited shifts consistent with greater rates of critical illness, including lower albumin (ALB) and higher glucose (GLU) variability (Supplementary Table~\ref{tab:norma_design}).

\subsection{Lab values are individualized}\label{lab-values-are-individualized}

Across all three cohorts, within-person variability was substantially smaller than between-person variability for most analytes, confirming that laboratory values fluctuate around narrow individual setpoints (Fig.~\ref{fig:figure2}b; Supplementary Table~\ref{tab:variability}). We quantified this finding using the individuality index (II), the ratio of within-person to between-person coefficient of variation, where values below 0.6 indicate high biological individuality.

Hematologic indices showed the lowest intra-individual variability relative to inter-individual variability. Mean corpuscular hemoglobin (MCH), mean corpuscular volume (MCV), and red blood cell count (RBC) had individuality indices ranging from approximately 0.2 to 0.6 across CHS, eICU-CRD, and INSPIRE, with values in eICU-CRD generally equal to or lower than those observed in CHS and INSPIRE (Supplementary Table~\ref{tab:variability}). In contrast, electrolytes and metabolic analytes exhibited lower biological individuality; potassium (K), glucose (GLU), sodium (NA), and calcium (CA) had individuality indices ranging from approximately 0.7 to 1.2 across cohorts, with within-person variability approaching the population spread (Fig.~\ref{fig:figure2}b; Supplementary Table~\ref{tab:variability}). INSPIRE showed patterns consistent with an intermediate regime between outpatient and ICU settings.

Within-person stability carried prognostic signal. Greater deviation from a patient's baseline, quantified as the absolute z-score of each index measurement relative to that patient's baseline average, was monotonically associated with mortality across all cohorts (Fig.~\ref{fig:figure2}c; Supplementary Fig.~\ref{fig:mortality_deviation}). When stratified by raw analyte values, several biomarkers, including white blood cell count (WBC), glucose (GLU), aspartate aminotransferase (AST), and hemoglobin A1c (HbA1c), exhibited U-shaped mortality curves, in which both abnormally high and low values were associated with elevated risk.

\subsection{NORMA generates calibrated individualized intervals}\label{norma-generates-calibrated-individualized-intervals}

NORMA is a conditional, autoregressive transformer framework with multiple configurations that models the distribution of a patient's next laboratory value given their longitudinal measurement history (Fig.~\ref{fig:figure3}a; Methods) and population data on ``healthy'' variation. The model is trained to forecast the next observed value. To derive a personalized reference interval, we condition the query token on values within the population reference range, so that the resulting 95\% prediction interval represents the expected range for that patient given their prior trajectory and population context, rather than a strictly disease-free state. We trained NORMA on 3.4 million longitudinal sequences from MIMIC-IV and EHRSHOT under two output parameterizations (Gaussian and quantile), which share the same architecture but differ in distributional assumptions (Methods and Supplementary Table~\ref{tab:cohort}).

On a held-out test set, NORMA outperformed all baselines in next-step forecasting accuracy across both output parameterizations. Baseline models included last value carried forward, autoregressive integrated moving average (ARIMA), and patient-specific mean. Mean absolute error was lowest for NORMA quantile (5.9 [IQR 3.6--9.1]) and NORMA Gaussian (6.0 [2.6--10.2]), compared with ARIMA (6.7 [4.3--11.9]), last value carried forward (7.3 [3.9--10.7]), and patient-specific mean (8.4 [4.5--13.6]; Supplementary Table~\ref{tab:prediction_performance}; Fig.~\ref{fig:figure3}b). Forecasting accuracy varied by analyte, with mean platelet volume (MPV) showing the weakest performance, consistent with its having the smallest training sample size (Fig.~\ref{fig:figure3}c; Supplementary Fig.~\ref{fig:analyte_heatmap}).

NORMA's prediction intervals responded appropriately to the factors that govern uncertainty (Fig.~\ref{fig:figure3}d; Supplementary Table~\ref{tab:sensitivity_summary}). More prior measurements narrowed intervals, longer prediction horizons widened them, and greater within-person variability produced substantially wider intervals. The quantile parameterization was more responsive than the Gaussian to changes in input features; for example, doubling within-person variability roughly doubled the predicted interval width under the quantile model, compared with minimal change under the Gaussian. Importantly, interval width stabilized after approximately 30 prior measurements and did not continue to narrow with additional data, indicating that the model does not overweight longer histories.

\subsection{NORMA$_{RI}$ detects abnormalities earlier than population intervals}\label{normari-detects-abnormalities-earlier-than-population-intervals}

In all three external validation cohorts, Per$_{RI}$ flagged the most tests as abnormal, Pop$_{RI}$ the fewest, and NORMA$_{RI}$ was intermediate. In CHS, abnormality rates were 29.6\% (Pop$_{RI}$), 39.1\% (NORMA$_{RI}$), and 46.8\% (Per$_{RI}$). In eICU-CRD, where baseline abnormality rates were higher due to acute illness, the same ordering held: 50.2\% (Pop$_{RI}$), 55.8\% (NORMA$_{RI}$), and 68.1\% (Per$_{RI}$). In INSPIRE, abnormality rates followed the same pattern, with values of 35.1\% (Pop$_{RI}$), 42.5\% (NORMA$_{RI}$), and 57.0\% (Per$_{RI}$), consistent with its intermediate clinical setting (Fig.~\ref{fig:figure4}a; Fig.~\ref{fig:figure5}a; Fig.~\ref{fig:figure6}a; Supplementary Table~\ref{tab:prevalence}).

Among tests classified as normal by population intervals, NORMA$_{RI}$ reclassified 12, 9, and 14 per 100 as abnormal in CHS, eICU-CRD, and INSPIRE, respectively, compared with 27, 37, and 34 per 100 for Per$_{RI}$ (Supplementary Tables~\ref{tab:prevalence}--\ref{tab:prevalence_interpret}).

NORMA$_{RI}$ flagged abnormalities earlier than population intervals across all cohorts. In CHS, 49\% of tests NORMA$_{RI}$ flagged as abnormal were later also flagged by Pop$_{RI}$, with a median lead time of 8.7 months (IQR 1.7--33.8; Supplementary Table~\ref{tab:lead_time}; Fig.~\ref{fig:figure4}e). In eICU-CRD, 25\% were later confirmed by Pop$_{RI}$, with a median lead time of 34.8 hours (IQR 23.1--72.6; Supplementary Table~\ref{tab:lead_time}; Fig.~\ref{fig:figure5}e). In INSPIRE, 30\% were later confirmed by Pop$_{RI}$, with a median lead time of 23.7 hours (IQR 11.4--54.7), consistent with its perioperative timescale (Fig.~\ref{fig:figure6}e; Supplementary Table~\ref{tab:lead_time}).

\subsection{NORMA$_{RI}$ detects clinically meaningful abnormalities missed by population intervals}\label{normari-detects-clinically-meaningful-abnormalities-missed-by-population-intervals}

We next asked whether the additional abnormalities detected by NORMA$_{RI}$ carried clinical meaning. Restricting to measurements classified as normal by population intervals, where personalization could add signal beyond standard practice, NORMA$_{RI}$ reclassifications carried substantially higher positive predictive value than Per$_{RI}$. For every 100 patients Pop$_{RI}$ classified as normal but NORMA$_{RI}$ flagged as abnormal in eICU-CRD, 13 died in hospital versus 10 under Per$_{RI}$ ($\Delta$ = +3), 16 developed AKI versus 15 ($\Delta$ = +1), and 27 had prolonged ICU stays versus 23 ($\Delta$ = +4). In CHS, 30 died versus 24 under Per$_{RI}$ ($\Delta$ = +6), 46 developed CKD versus 39 ($\Delta$ = +7), and 80 had type 2 diabetes versus 75 ($\Delta$ = +5). In INSPIRE, NORMA$_{RI}$ identified 6 more unplanned ICU admissions, 1 more in-hospital death, and 1 more perioperative infection per 100 reclassified patients than Per$_{RI}$ (Supplementary Tables~\ref{tab:chs_eval_interpret}--\ref{tab:inspire_eval_interpret}). Across all assessed outcomes -- mortality, type 2 diabetes, and CKD over 10 years in CHS; in-hospital mortality, AKI, sepsis, and prolonged LOS in eICU-CRD; and perioperative mortality, prolonged LOS, unplanned ICU admission, and postoperative infection in INSPIRE -- Per$_{RI}$ achieved higher sensitivity but NORMA$_{RI}$ achieved substantially higher specificity and positive predictive value (Figs.~\ref{fig:figure4}c,d, \ref{fig:figure5}c,d, \ref{fig:figure6}c,d and Supplementary Figs.~\ref{fig:eval_dotplot_chs}--\ref{fig:eval_dotplot_inspire}).

\subsection{NORMA$_{RI}$ improves clinical prediction}\label{normari-improves-clinical-prediction}

In time-to-event analyses across all cohorts, NORMA$_{RI}$ and Pop$_{RI}$ abnormality flags showed comparable prognostic associations with clinical outcomes, while Per$_{RI}$ flags were substantially weaker (Figs.~\ref{fig:figure4}b, \ref{fig:figure5}b, \ref{fig:figure6}b and Supplementary Tables~\ref{tab:chs_cox_mortality}--\ref{tab:inspire_cox_unplanned_icu}). This was most pronounced in CHS, where NORMA$_{RI}$ abnormality flags were strongly associated with all-cause mortality for hematologic and metabolic analytes, with several showing 2- to 4-fold elevated risk after adjusting for age and sex -- including creatinine (HR 3.30), hemoglobin (HR 3.72), and hematocrit (HR 3.36). Effect sizes were comparable to or slightly exceeded those of Pop$_{RI}$ for most analytes, while Per$_{RI}$ flags carried essentially no prognostic signal.

Mean concordance indices for mortality were similar between NORMA$_{RI}$ and Pop$_{RI}$ across all three cohorts (CHS: 0.752 vs. 0.758; eICU-CRD: 0.567 vs. 0.571; INSPIRE: 0.730 vs. 0.731), while Per$_{RI}$ was consistently lowest (CHS: 0.748; eICU-CRD: 0.562; INSPIRE: 0.695). The gap was most pronounced in INSPIRE, where Per$_{RI}$ concordance (median 0.718, IQR [0.673, 0.742]) fell well below both NORMA$_{RI}$ (0.754 [0.712, 0.780]) and Pop$_{RI}$ (0.742 [0.722, 0.778]; Supplementary Tables~\ref{tab:chs_cox_mortality}, \ref{tab:eicu_cox_mortality}, \ref{tab:inspire_cox_mortality}).

\section{Discussion}
Routine laboratory testing is becoming more frequent, more accessible, and more central to medical decision-making\cite{Koch2018-jj,Jorgensen2022-xk}. Beyond tests that clinicians order within the healthcare system, patients are increasingly seeking out repeated panel-based screening through direct-to-consumer companies. In parallel, interpretation is shifting away from static cutoffs and monolithic ``normal ranges'' towards ``personalized'' ranges that leverage trajectories of variation and individual baselines, yet the implications and risks of this shift remain largely unclear\cite{Coskun2022-pp,Tayob2022-vz,Coskun2021-xa,Wang2022-qg}. Across nearly two billion laboratory measurements spanning outpatient and intensive care settings, we find that purely personalized reference intervals consistently overcall abnormalities and are poorly associated with adverse outcomes. NORMA balances purely personalized and population intervals by training a transformer to predict what a patient's next laboratory value is expected to fall if they remain physiologically stable, and defining the reference interval from this conditional distribution. In doing so, NORMA$_{RI}$ inherits the sensitivity of personalized interpretation while maintaining the specificity of population-level definitions of normal variation.

In eICU, Per$_{RI}$ flagged the majority of laboratory measurements as abnormal, and similarly elevated abnormality rates were observed in CHS and INSPIRE compared to population-based intervals, indicating that a purely personalized framework would likely generate alerts for a substantial fraction of tests across critical care, outpatient, and perioperative settings. At the scale of a national health system processing millions of tests annually, this could translate to an enormous burden of false alarms, unnecessary follow-up testing, and potential patient anxiety\cite{Kohane2006-tq,Naugler2018-mf,Badrick2025-rn}. By contrast, NORMA$_{RI}$ moderated this alert burden, flagging fewer measurements than Per$_{RI}$ while yielding abnormalities that were more strongly enriched for adverse clinical outcomes.

This benefit was not apparent from aggregate discrimination alone. Across individual analytes and outcomes, NORMA$_{RI}$ and Pop$_{RI}$ achieved broadly similar concordance indices, whereas Per$_{RI}$ consistently performed worst. This similarity between NORMA$_{RI}$ and Pop$_{RI}$ might suggest that personalization adds little value; however, the value of NORMA$_{RI}$ is seen in individuals that Pop$_{RI}$ labels normal but NORMA$_{RI}$ flags as abnormal. In this reclassified subset, NORMA$_{RI}$ achieved substantially higher precision and specificity than Per$_{RI}$.

By anchoring individualized intervals to population-level expectations, NORMA reduces the risk that chronic or subclinical disease is absorbed into a patient's estimated baseline, a limitation of purely personalized approaches such as Per$_{RI}$. This limitation is not unique to the Gaussian mixture approach used for Per$_{RI}$. Even with Bayesian updating or hierarchical extensions, a population prior converges toward the patient's observed distribution as measurements accumulate, eventually washing out the prior and reproducing the same overcalling behavior\cite{Roshan2021-tj,Tayob2022-vz}. However, such methods are also nondeterministic, computationally expensive, introduce prior-strength and group-structure hyperparameters, and lack native handling of time or irregular sampling. NORMA avoids this by conditioning on a healthy state at every prediction, regardless of how many measurements are available, which is why its interval width stabilizes rather than continuing to narrow. Furthermore, it does not require explicit pre-filtering of trajectories for stability; it handles irregular measurement spacing natively and its inputs are limited to data already present in most electronic health records. In a clinical deployment, NORMA could provide more "personalized" assessment for patients even without sufficient measurement history.

NORMA builds on a parallel line of work that scales sequence modelling over longitudinal health records into clinical foundation models. Early efforts demonstrated that deep learning over raw EHR streams could match or exceed task-specific models for mortality, readmission, and laboratory forecasting\cite{Rajkomar2018-hg,McDermott2023-xd,Yang2023-ei,Jiang2023-iq}. More recent transformer-based foundation models, trained on event streams from millions of patient timelines, have demonstrated that longitudinal records can support zero- or few- shot forecasting of diagnoses, procedures, and disease progression at increasing scale, with prediction horizons extending years into the future and architectures expanding to incorporate multimodal clinical data\cite{Shmatko2025-rg,Waxler2025-gu,Renc2024-so,Zhang2026-dt}. The same paradigm is now extending beyond structured EHRs to continuous physiological streams such as continuous glucose monitoring and other wearables\cite{Lutsker2026-xg,Metwally2026-qy}.

These models demonstrate that longitudinal patient representations can support broad outcome prediction across clinical domains. While NORMA learns from patient trajectories, it is distinct in the prediction task. Rather than learning a general representation of trajectories, such as APOLLO, to predict diagnoses, procedures, or downstream clinical outcomes, it models continuous biomarker distributions and estimates the range of values expected for a given patient under a specified future laboratory state\cite{Zhang2026-dt}. NORMA therefore draws on counterfactual prediction and sequence-editing frameworks, such as CLEF, which ask how a trajectory would change under an imposed condition or intervention\cite{Melnychuk2022-rq,Li2025-ff,Isaac2025-ef}. Here, the imposed condition is a future normal laboratory state, allowing NORMA to estimate an individualized healthy-state reference interval rather than simply forecast the next observed measurement. The clinical outcomes evaluated here, including all-cause mortality, acute kidney injury, type 2 diabetes, and chronic kidney disease, were used only for downstream validation rather than model training. This disease-agnostic design suggests that deviations from an individual's expected healthy trajectory may carry prognostic relevance across diverse disease states without requiring outcome-specific retraining. As longitudinal biomarker monitoring expands further into precision medicine, the challenge of reconciling population-derived and individualized reference standards will broaden, and conditional prediction anchored to a healthy reference state may generalize across these settings\cite{Isaac2025-ef,Lutsker2026-xg,Metwally2026-qy,McIntosh2002-op}.

Our study had several limitations. First, NORMA's current scope is limited by the analytes and clinical inputs used for training. NORMA was trained and evaluated on 30 common blood tests and conditioned only on age, sex, and laboratory trajectories; it does not currently incorporate comorbidities, medications, or other clinical context, which could further refine expected trajectories and abnormality thresholds\cite{Johnson2025-ay}. Performance was also not uniform across analytes. MPV showed the lowest forecasting accuracy, consistent with its having the smallest training sample size and weakest forecasting accuracy among the 30 analytes. Second, some modeling assumptions may affect calibration. In CHS, we modeled predictive distributions as Gaussian, which may not fully capture skewed or heavy-tailed analytes. However, we implemented quantile regression in eICU, and results were consistent across both parameterizations, suggesting that NORMA's clinical value does not depend on a strict Gaussian assumption. Broader exploration of output distributions may further improve calibration for skewed or heavy-tailed analytes. Third, generalizability and clinical impact remain to be established prospectively. The CHS analysis was conducted within a single national health system; eICU, although geographically diverse, represents only the ICU segment of hospital care; and INSPIRE reflects perioperative care at a single academic medical center. Performance may differ in emergency departments, primary care clinics, non-surgical inpatient settings, or populations with different demographics and laboratory utilization patterns. Finally, this study evaluated retrospective associations between abnormality flags and clinical outcomes. Prospective trials are needed to determine whether NORMA$_{RI}$-augmented interpretation accelerates time to diagnosis, alters downstream testing, and influences clinician behavior. Such evaluation may further benefit from cohort-specific fine-tuning, for example on Clalit data before deployment, which could improve local performance at the cost of generalizability.

Individualized laboratory interpretation can be made more precise and more clinically useful by combining patient-specific trajectories with population-level expectations and uncertainty-aware prediction. By validating across longitudinal outpatient and acute settings, outcome horizons, and patient populations, we show that this framework generalizes beyond a single clinical context. NORMA provides a framework for integrating personalized laboratory interpretation into routine practice, balancing the sensitivity of personalized intervals with the specificity of population-anchored prediction.

\section{Methods}\label{methods}

\subsection{Study Populations}\label{study-populations-1}

We used five cohorts spanning model development and external validation. Model development used MIMIC-IV and EHRSHOT. External validation used three independent cohorts representing distinct clinical settings and time horizons: Clalit Health Services (CHS), a national longitudinal health system cohort from Israel; the eICU Collaborative Research Database (eICU-CRD), a multicenter critical care cohort from the United States; and INSPIRE, a perioperative cohort from Seoul National University Hospital in South Korea. Across these cohorts, we evaluated three reference interval frameworks: population-based reference intervals (PopRI), personalized reference intervals (Per$_{RI}$), and NORMA-derived reference intervals (NORMA$_{RI}$). No development data were used in external validation.

\paragraph{MIMIC-IV and EHRSHOT}
MIMIC-IV contributed 179,601 patients from a tertiary-care hospital system and was used for acute-care, short-horizon model development. EHRSHOT contributed 5,676 patients from longitudinal electronic health records and was used for intermediate-horizon model development. Together, these cohorts provided the development data used for NORMA training, validation, and held-out testing.

\paragraph{Clalit Health Services}
The CHS cohort comprises nationwide longitudinal electronic healthcare data from Israel's largest health maintenance organization\cite{Balicer2011-ud}. We included adults aged 18--99 years with repeated outpatient laboratory testing between 2000 and 2024. For each analyte, eligibility required at least five outpatient measurements spaced at least 90 days apart prior to January 1, 2015 (baseline period) and at least one measurement between January 1, 2015 and January 1, 2016 (classification period). We excluded hospitalizations, emergency department visits, and urgent encounters during baseline ascertainment. The first measurement during the classification period served as the index value and defined the index date. We assessed eligibility independently per analyte, allowing individuals to contribute to multiple analyses.

We followed individuals for up to ten years from each analyte-specific index date. For those without events, we censored follow-up at the earliest of health plan disenrollment or death. Patients remaining enrolled without a recorded death were censored at the last available date in the dataset (January 1, 2024). Primary outcomes included all-cause mortality and incident diagnoses of type 2 diabetes (T2D) and chronic kidney disease (CKD). We ascertained outcomes using ICD-9/10 codes supplemented by laboratory criteria. We defined T2D as HbA1c $\geq$6.5\% or fasting glucose $\geq$126 mg/dL on two occasions; CKD as an estimated glomerular filtration rate <60 mL/min/1.73 m² on two measurements at least 90 days apart.

\paragraph{eICU Collaborative Research Database}
The eICU Collaborative Research Database (eICU-CRD v2.0) is a multicenter critical care cohort comprising more than 200,859 ICU admissions from 208 hospitals across the United States\cite{Pollard2018-ru}. We included adults aged 18--99 years with repeated laboratory measurements during their ICU stay. We applied a baseline filter requiring at least five measurements. After filtering, 98,432 unique patients contributed over 20 million measurements across 29 analytes.

For each patient--analyte sequence, we used the first 75\% of measurements as the baseline to estimate Per$_{RI}$ and NORMA$_{RI}$, and we used the remaining measurements as index values for classification. Primary outcomes included in-hospital mortality, acute kidney injury (AKI), sepsis, and prolonged ICU length of stay (>7 days). Outcomes were ascertained using ICD-9 diagnosis codes and structured clinical documentation.

\paragraph{INSPIRE}
The INSPIRE dataset is a publicly available perioperative research cohort comprising approximately 130,000 surgical cases from a single academic medical center (Seoul National University Hospital) in South Korea between 2011 and 2020\cite{Lim2024-rl}. We included adults aged 20--90 years with repeated laboratory measurements during the perioperative period. We applied a baseline filter requiring at least five measurements. After filtering, 51,159 unique patients contributed over 10 million measurements across 19 analytes.

For each patient--analyte sequence, we used the first 75\% of measurements as the baseline to estimate Per$_{RI}$ and NORMA$_{RI}$, and we used the remaining measurements as index values for classification. Primary outcomes included in-hospital mortality, prolonged length of stay, unplanned ICU admission, and perioperative infection. Outcomes were ascertained using ICD-10-CM diagnosis codes and structured clinical documentation.

\subsection{Laboratory Measurements}\label{laboratory-measurements}

We selected 30 routinely measured blood analytes from common clinical panels based on clinical ubiquity and sufficient repeated measurement across development and validation datasets. We grouped analytes into four standard clinical panels: complete blood count (hematocrit, hemoglobin, mean corpuscular hemoglobin, mean corpuscular hemoglobin concentration, mean corpuscular volume, mean platelet volume, platelet count, red blood cell count, red cell distribution width, white blood cell count); comprehensive metabolic panel (sodium, potassium, chloride, calcium, bicarbonate, blood urea nitrogen, creatinine, glucose, aspartate aminotransferase, alanine aminotransferase, alkaline phosphatase, total bilirubin, direct bilirubin, total protein, albumin); lipid panel (total cholesterol, low-density lipoprotein cholesterol, high-density lipoprotein cholesterol, triglycerides); glycemic control (glycated hemoglobin).

We mapped all laboratory results to Logical Observation Identifiers Names and Codes (LOINC)\cite{McDonald2003-px}. We retained only positive values, standardized units to canonical formats, and excluded analyte-specific outliers exceeding ±3 standard deviations from the analyte median. When duplicate results shared identical timestamps, we averaged values within each analyte.

\subsection{Reference Interval Frameworks}\label{reference-interval-frameworks}

\paragraph{Population Reference Intervals}
We defined Pop$_{RI}$ using externally specified laboratory reference ranges and clinical decision thresholds rather than estimating cohort-specific intervals. For each analyte, we used the adult reference range reported by the American Board of Internal Medicine Laboratory Test Reference Ranges\cite{American_Board_of_Internal_Medicine2025-jc}. Where sex-specific reference intervals were provided, classifications were made using the patient's sex-specific range. Supplementary Table~\ref{tab:analyte_reference} provides the reference interval, sex stratification, and canonical unit used for each analyte.

\paragraph{Personalized Reference Intervals}
We estimated Per$_{RI}$ following Foy \emph{et al.} For each patient-analyte pair, we fit Gaussian mixture models with one to three components to baseline measurements and selected model order using the Akaike information criterion\cite{Foy2025-dj}. From the selected model, we chose the mixture component that explained the largest number of that patient's baseline measurements and treated it as the individual's physiological setpoint. We defined the Per$_{RI}$ as the mean of this component ± 2 standard deviations.

\subsection{NORMA (Normal Outcome Range Modeling with Attention)}
NORMA is a conditional, decoder-only transformer that models the distribution of a patient's next laboratory value given their longitudinal measurement history and a query specifying a future health state and prediction horizon. We trained the model on 3.4 million longitudinal sequences from MIMIC-IV and EHRSHOT, requiring at least three repeated measurements per patient without constraints on sampling intervals or clinical setting\cite{Johnson2024-ls,Wornow2023-fd}. Data were split at the patient level into training (70\%), validation (10\%), and test (20\%) sets to prevent leakage across partitions. Each sequence was tokenized, padded, and masked to handle variable-length histories. A weighted sampling scheme was used during training to balance representation across 30 laboratory analytes. Alternative input encodings with justifications are summarized (Supplementary Table~\ref{tab:norma_design}).

\paragraph{Input representation}
For a given patient-biomarker pair, we denote the ordered sequence of observed laboratory values as x = (x$_{1}$, \ldots, x$_{T}$), with corresponding measurement times t = (t$_{1}$, \ldots, t$_{T}$) and clinical states s = (s$_{1}$, \ldots, s$_{T}$), where each s$_{i}$ indicates the laboratory state relative to the population reference interval (low, normal, or high).

The model processes these inputs as a sequence of three token types: a context token, history tokens, and a query token. The context token $z_{c}$ encodes static patient covariates: sex \emph{g}, age \emph{a}, and laboratory test code \emph{c}, as a sum of learned embeddings:

\begin{equation*}
z_{c} = e_{g}(g) + e_{a}(a) + e_{c}(c)
\end{equation*}

Each history token $z_{i}^{hist}$ represents one prior measurement and combines three components:

\begin{equation*}
z_{i}^{hist} = f_{v}(\tilde{x}_{i}) + e_{s}(s_{i}) + f_{\tau}(\Delta t_{i})
\end{equation*}

where $f_{v}$ is a linear projection of the input value $\tilde{x}_{i}$, $e_{s}$ is a learned state embedding, and $f_{\tau}$ encodes the inter-measurement interval $\Delta t_{i} = t_{i} - t_{i-1}$.

The query token, $z_{query},$ specifies the context at the time of prediction:

\begin{equation*}
z_{query} = e_{s}(s_{T+1}) + f_{h}(t_{T+1} - t_{T})
\end{equation*}

where $s_{T+1}$ denotes the future health state and $t_{T+1} - t_{T}$ the prediction horizon. The full sequence is processed using a causally masked Transformer decoder to obtain the predictive distribution of the next observed laboratory value by $P(x_{T+1} | x, t, s, g, a, c, s*, t*).$

\paragraph{Training Objective}
Given the input sequence, the model is optimized to predict the conditional distribution of the next observed value x$_{T+1}$. We trained NORMA under two output parameterizations.

Under the Gaussian parameterization, the model jointly estimates a predicted mean $\mu$ and log-variance log $\sigma$²:

\begin{equation*}
L = \tfrac{1}{2} [\log \sigma^{2} + (x_{T+1} - \mu)^{2} / \sigma^{2}]
\end{equation*}

Under the quantile parameterization, the model directly predicts fixed quantiles $\tau \in$ \{0.025, 0.25, 0.50, 0.75, 0.975\} using a pinball loss without distributional assumptions:

\begin{equation*}
L_{\tau} = \tau \times \max(0, x_{T+1} - q_{\tau}) + (1-\tau) \times \max(0, q_{\tau} - x_{T+1})
\end{equation*}

where q$_{\tau}$ is the predicted $\tau$-th quantile. Both parameterizations share the same core architecture and were trained on the same development data.

\paragraph{NORMA reference intervals}
We define the NORMA reference interval (NORMA$_{RI}$) as the 95\% prediction interval obtained by conditioning on a future normal laboratory state. Under the Gaussian parameterization, this corresponds to [$\mu -$ 1.96$\sigma$, $\mu$ + 1.96$\sigma$]; under the quantile parameterization, it corresponds to [$\hat{q} _{0.025}$, $\hat{q} _{0.975}$].

\paragraph{Evaluation}
We first evaluated forecasting performance for next-step prediction. For point accuracy, we used the 50th percentile (median) prediction and compared it against three baselines: a person-specific historical mean model, last observation carried forward ("Last"), and an autoregressive integrated moving average model (ARIMA). Performance was quantified using mean absolute error (MAE), mean absolute percentage error (MAPE), and coefficient of determination (R²).

\paragraph{Sensitivity Analysis}
To assess how NORMA captures uncertainty across clinical contexts, we conducted one-at-a-time sensitivity sweeps over five input features: patient age, sex, number of prior measurements, prediction horizon, and within-sequence variability. For each of 30 laboratory tests, we constructed a synthetic baseline sequence: 10 measurements spaced 90 days apart, each set to the midpoint of the sex-specific population reference interval, for a 50-year-old male with a 30-day prediction horizon. We then varied one feature at a time while holding all others at baseline values. Age was swept from 20 to 80 years in 5-year increments; history length from 2 to 300 measurements; and prediction horizon from 7 days to 10 years. For within-sequence variability, we parameterized the standard deviation as a multiplier of one-tenth of the reference range width (0.0 to 3.0) and, at each noise level, drew 30 independent histories from N(midpoint, $\sigma$) to capture sampling variability. For each configuration, we recorded the predicted median (50th quantile), the 90\% prediction interval width, and the percent change in interval width relative to baseline, enabling direct comparison of sensitivity magnitude across biomarkers.

\paragraph{NORMA configuration}
For CHS validation, we applied NORMA with the default configuration: Time2Vec temporal encoding, ternary health-state indicators, raw-value laboratory inputs, raw age projected linearly, and the Gaussian output parameterization. For eICU validation, we used log-delta-t temporal encoding with periodic components, ternary health-state indicators, within-sequence normalization of laboratory values, binned age embeddings in decade-wide bins, a dedicated context token for patient covariates, and the quantile output parameterization. INSPIRE used the same configuration as eICU. All cohort-specific NORMA configurations are summarized in Supplementary Table~\ref{tab:norma_design}.

\subsection{Statistical Analysis}\label{statistical-analysis}

\paragraph{Mortality association analysis}
We assessed the association between laboratory values and mortality using two complementary approaches. First, we grouped index-period measurements into quintiles of raw analyte values within each analyte and estimated mortality rates per quintile. Second, to evaluate whether deviation from a patient's personal baseline predicts mortality, we computed a deviation score for each index measurement as the absolute z-score from the patient's baseline mean (|value $- \mu _{baseline}$| / $\sigma _{baseline}$), where $\mu _{baseline}$ and $\sigma _{baseline}$ are the mean and standard deviation of the patient's baseline measurements. We grouped deviation scores into deciles and estimated mortality rates within each decile. For both analyses, we computed 95\% confidence intervals using Wilson score intervals. Analytes with fewer than 20 observations were excluded.

\paragraph{Abnormality classification}
We classified each index measurement as normal or abnormal under Pop$_{RI}$, Per$_{RI}$, and NORMA$_{RI}$. To avoid incorporating pre-existing pathology into Per$_{RI}$ thresholds, we excluded patients whose Per$_{RI}$ mean fell outside the Pop$_{RI}$. Measurements classified as Pop$_{RI}$-abnormal were also classified as abnormal using Per$_{RI}$ and NORMA$_{RI}$.

\paragraph{Clinical Risk Stratification and Prediction}
We evaluated whether abnormal classifications under each framework identified individuals at elevated risk of future events. For each analyte and framework (Pop$_{RI}$, Per$_{RI}$, NORMA$_{RI}$), we defined a binary indicator of abnormality for values outside the corresponding interval.

For each analyte--outcome pair, we calculated positive predictive value (probability of the event given abnormal classification), sensitivity (probability of abnormal classification among those with the event), and specificity (probability of normal classification among those without the event). To isolate differences between interval paradigms, we restricted analyses to measurements within the Pop$_{RI}$, because values outside the Pop$_{RI}$ necessarily fall outside Per$_{RI}$ and NORMA$_{RI}$.

We estimated associations using Cox proportional hazards models with a single binary abnormality indicator adjusted for age and sex. We split data into training (60 percent) and test (40 percent) sets stratified by event status and approximate event time in one-year bins. We evaluated performance at 1, 3, 5, and 10 years using time-dependent receiver operating characteristic curves and quantified uncertainty with 95\% confidence intervals from 1,000 bootstrap resamples. We applied Benjamini-Hochberg false discovery rate correction to account for multiple comparisons across outcomes and analytes.

\section{Ethics Approval}\label{ethics-approval}

The use of Clalit Health Services data in this study was approved by the Clalit Health Services Institutional Review Board (Helsinki Committee). MIMIC-IV, eICU-CRD, and INSPIRE were accessed under the PhysioNet Credentialed Health Data Use Agreement; EHRSHOT was accessed under the Stanford Research Use Agreement. All datasets were de-identified prior to release.

\section{Data Availability}\label{data-availability}

This study utilizes two publicly available datasets for the development of NORMA: EHRSHOT and MIMIC-IV, both accessible to qualified researchers under their respective data-use agreements. The eICU Collaborative Research Database is also publicly available via PhysioNet. Clalit Health Services data are not publicly available.

\section{Code Availability}\label{code-availability}

The full source code for the NORMA model, including data processing, training, and evaluation pipelines, is publicly available at \href{https://github.com/aashnapshah/NORMA}{\underline{https://github.com/aashnapshah/NORMA}}. The interactive web-based user interface for individualized laboratory interpretation is available at \href{https://norma-tpy0.onrender.com/}{\underline{https://norma-tpy0.onrender.com/}}.

\section{Acknowledgments}\label{acknowledgments}

We gratefully acknowledge support from the Ivan and Francesca Berkowitz Family Living Laboratory Collaboration at Harvard Medical School and Clalit Research Institute, NIEHS R01ES032470, and NIDDK R01DK137993.

\clearpage
\bibliographystyle{unsrtnat}
\bibliography{references}
\clearpage

\section*{Figures}
\renewcommand{\figurename}{Fig.}
\renewcommand{\thefigure}{\arabic{figure}}

\begin{figure}[H]
\centering
\includegraphics[width=\textwidth,height=0.78\textheight,keepaspectratio]{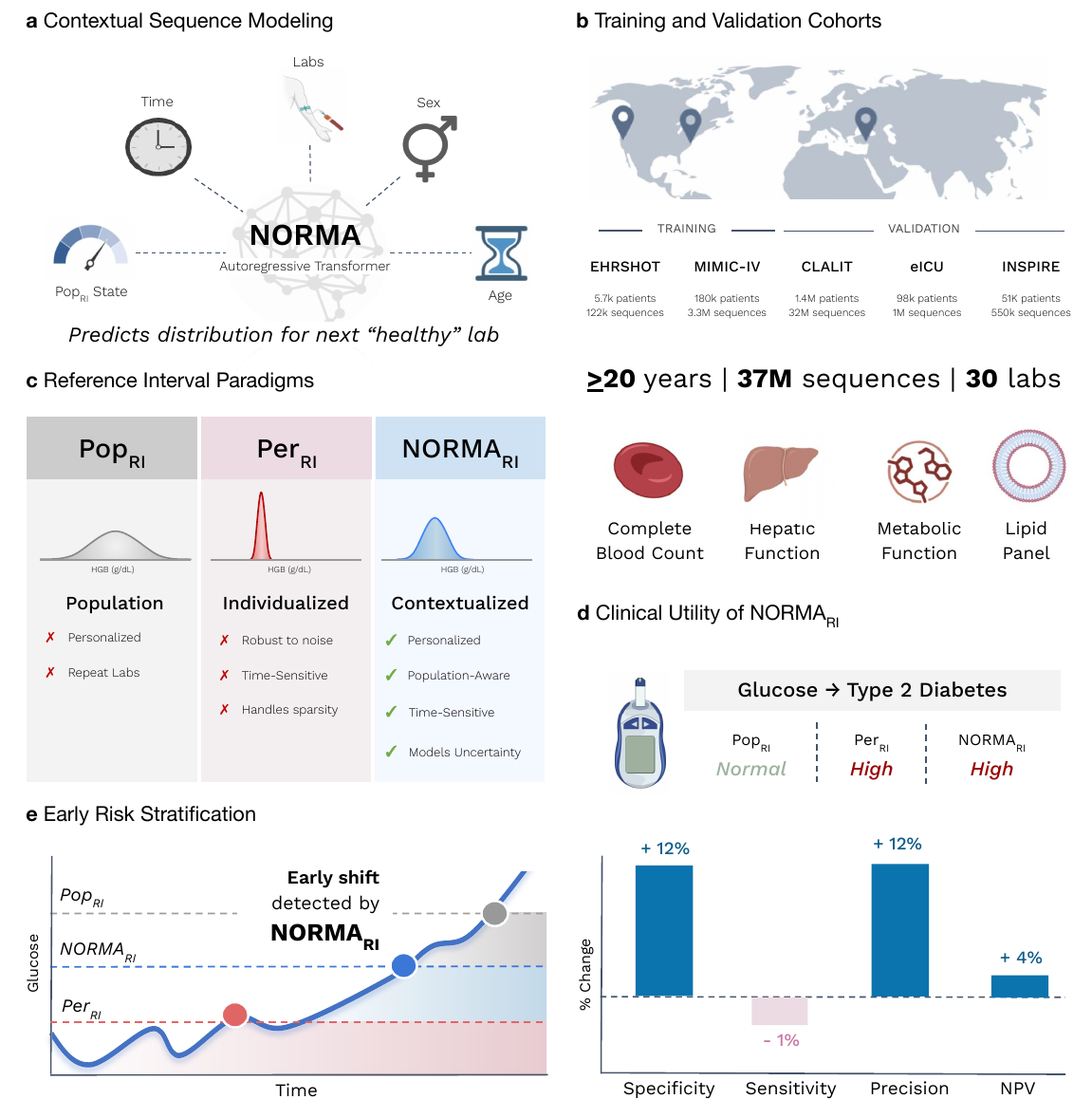}
\caption{\textbf{NORMA framework overview.} \textbf{a)} NORMA conditions on laboratory history, time, sex, age, and population reference state to predict the distribution of the next healthy laboratory value. \textbf{b)} Study design: training on EHRSHOT and MIMIC-IV; external validation on Clalit Health Services, the eICU Collaborative Research Database, and the INSPIRE cohort. Across all cohorts: 20 years, 37 million sequences, and 30 analytes spanning four clinical panels. \textbf{c)} Comparison of Pop$_{\mathrm{RI}}$, Per$_{\mathrm{RI}}$, and NORMA$_{\mathrm{RI}}$. \textbf{d)} Illustrative example: a glucose value classified as normal by Pop$_{\mathrm{RI}}$ but flagged abnormal by Per$_{\mathrm{RI}}$ and NORMA$_{\mathrm{RI}}$, with corresponding change in specificity, sensitivity, and precision for type 2 diabetes classification. \textbf{e)} Schematic of early abnormality detection by NORMA$_{\mathrm{RI}}$ relative to Pop$_{\mathrm{RI}}$ and Per$_{\mathrm{RI}}$ over time. Created with BioRender.com.}
\label{fig:figure1}
\vspace{2pt}
\end{figure}
\clearpage

\begin{figure}[H]
\centering
\includegraphics[width=\textwidth,height=0.78\textheight,keepaspectratio]{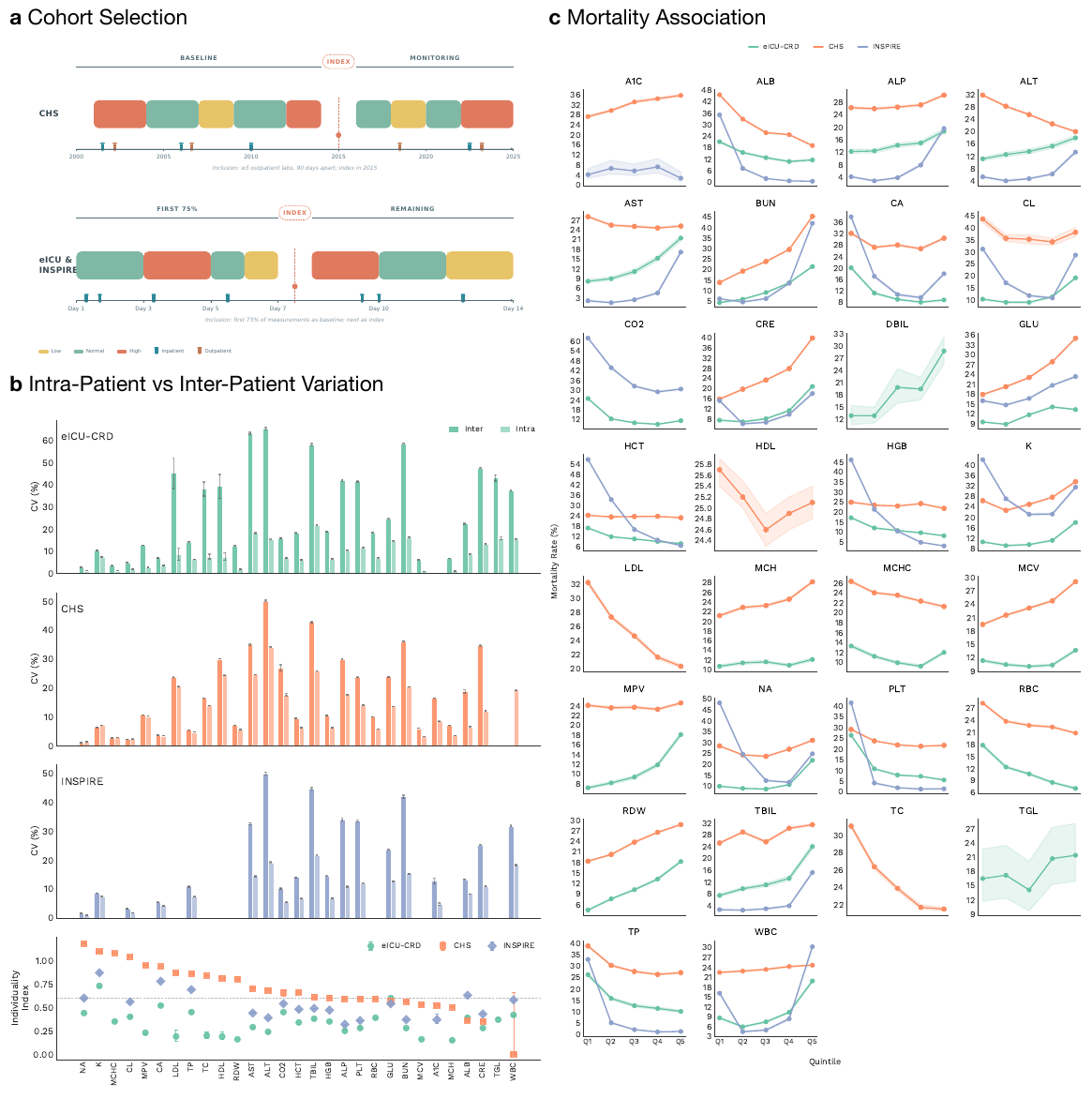}
\caption{\textbf{Within-person biological variability and mortality association.} \textbf{a)} Cohort Selection. Cohort selection and inclusion criteria for Clalit Health Services, the eICU Collaborative Research Database, and the INSPIRE cohort. \textbf{b)} Intra-Patient vs Inter-Patient Variation. Within-person versus between-person coefficient of variation for 30 analytes, with the individuality index shown below. The dashed line indicates an individuality index of 0.6. \textbf{c)} Mortality Association. Observed mortality rate by quintile of $z$-score from the patient baseline mean in Clalit Health Services, the eICU Collaborative Research Database, and the INSPIRE cohort. Shaded bands denote 95\% confidence intervals.}
\label{fig:figure2}
\vspace{2pt}
\end{figure}
\clearpage

\begin{figure}[H]
\centering
\includegraphics[width=\textwidth,height=0.78\textheight,keepaspectratio]{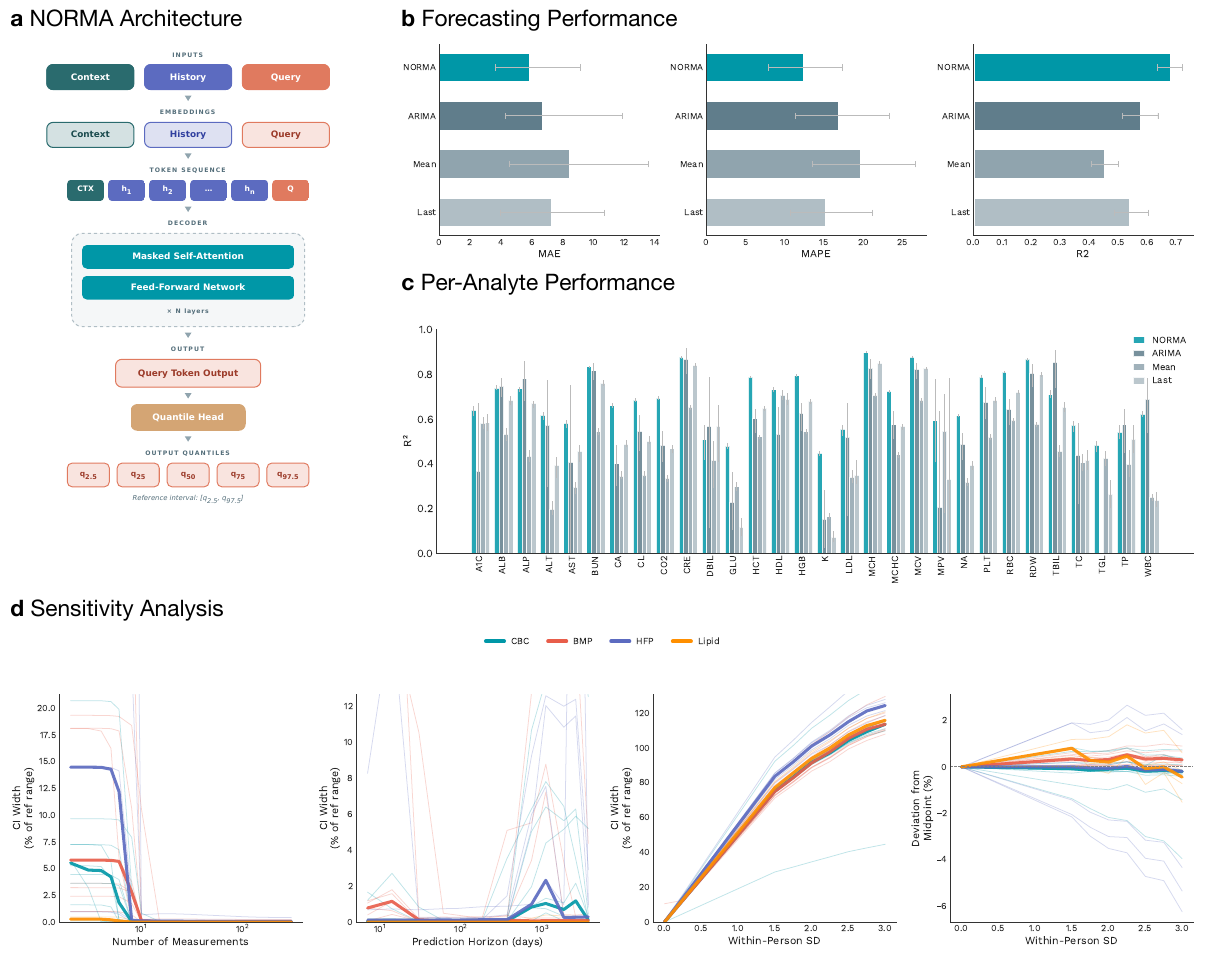}
\caption{\textbf{NORMA architecture, forecasting accuracy, and sensitivity analysis.} \textbf{a)} NORMA Architecture. Model architecture: demographic context, laboratory history, and query tokens are jointly embedded and processed by a causal transformer decoder to produce quantile predictions for the next laboratory value. \textbf{b)} Forecasting Performance. Held-out test-set forecasting accuracy (mean absolute error, mean absolute percentage error, and coefficient of determination) for NORMA (quantile) versus ARIMA, population mean, and last-value-carried-forward baselines. Error bars denote 95\% bootstrap confidence intervals. \textbf{c)} Per-Analyte Performance. Per-analyte coefficient of determination grouped by clinical panel. \textbf{d)} Sensitivity Analysis. Sensitivity of prediction interval width and midpoint deviation (each as a percentage of Pop$_{\mathrm{RI}}$ width) to history length, forecast horizon, and within-person variability.}
\label{fig:figure3}
\vspace{2pt}
\end{figure}
\clearpage

\begin{figure}[H]
\centering
\includegraphics[width=\textwidth,height=0.78\textheight,keepaspectratio]{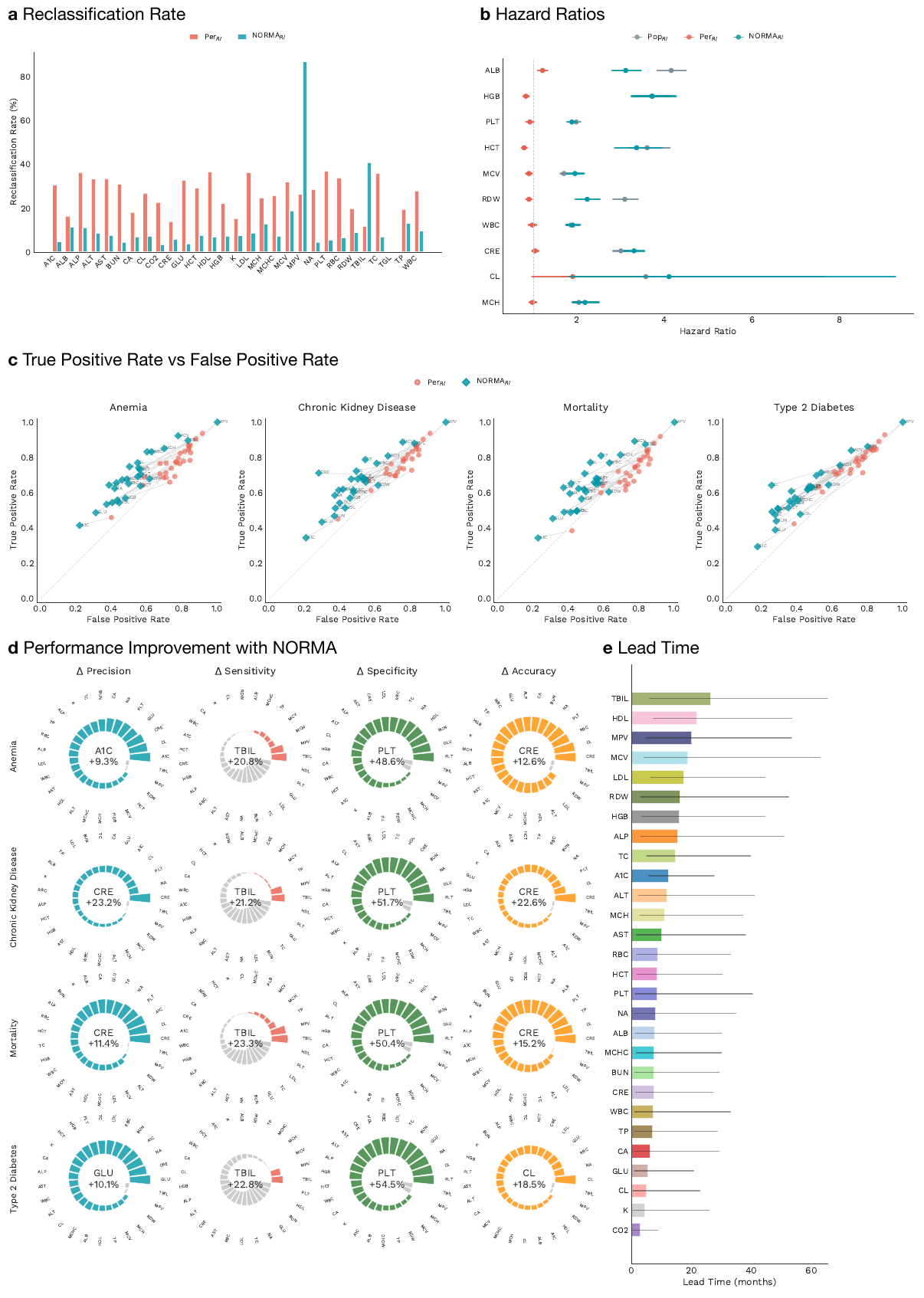}
\caption{\textbf{Reclassification performance and early risk stratification in Clalit Health Services.} \textbf{a)} Reclassification Rate. Reclassification rate by analyte among Pop$_{\mathrm{RI}}$-normal tests. \textbf{b)} Hazard Ratios. Cox hazard ratios for all-cause mortality (adjusted for age and sex); analytes with $p < 0.05$ shown. \textbf{c)} True Positive Rate vs False Positive Rate. True positive rate versus false positive rate for clinical outcome prediction across anemia, chronic kidney disease, all-cause mortality, and type 2 diabetes, comparing Per$_{\mathrm{RI}}$ and NORMA$_{\mathrm{RI}}$. \textbf{d)} Performance Improvement with NORMA. Change in precision, sensitivity, specificity, and balanced accuracy (NORMA$_{\mathrm{RI}}$ versus Per$_{\mathrm{RI}}$); inner label shows the largest gain. \textbf{e)} Lead Time. Median lead time from NORMA$_{\mathrm{RI}}$ flag to first Pop$_{\mathrm{RI}}$ flag, by analyte (months). Error bars denote the interquartile range.}
\label{fig:figure4}
\vspace{2pt}
\end{figure}
\clearpage

\begin{figure}[H]
\centering
\includegraphics[width=\textwidth,height=0.78\textheight,keepaspectratio]{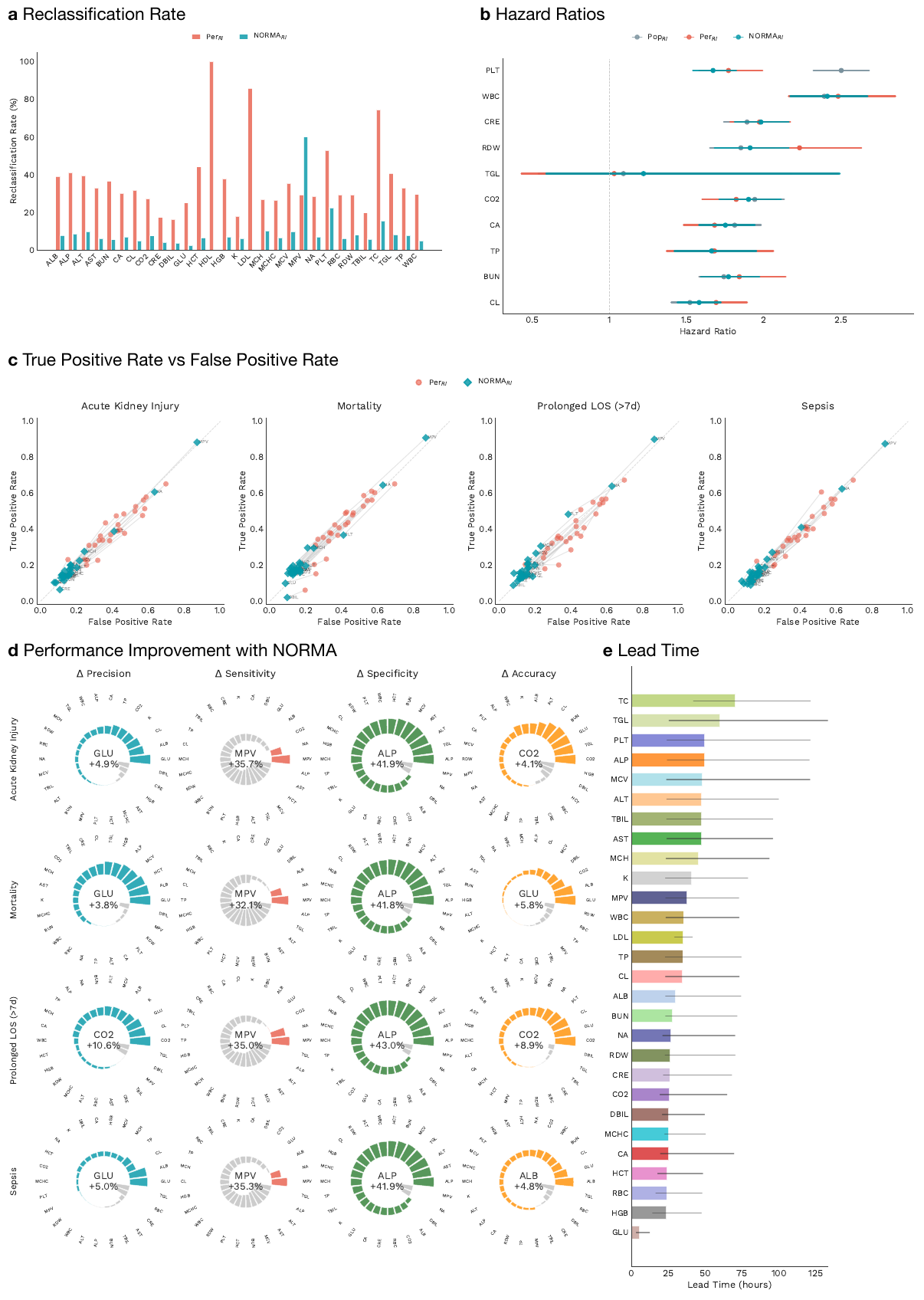}
\caption{\textbf{Reclassification performance and early risk stratification in the eICU Collaborative Research Database.} \textbf{a)} Reclassification Rate. Reclassification rate by analyte among Pop$_{\mathrm{RI}}$-normal tests. \textbf{b)} Hazard Ratios. Cox hazard ratios for in-hospital mortality (adjusted for age and sex); analytes with $p < 0.05$ shown. \textbf{c)} True Positive Rate vs False Positive Rate. True positive rate versus false positive rate for clinical outcome prediction across in-hospital mortality, acute kidney injury, sepsis, and prolonged ICU stay, comparing Per$_{\mathrm{RI}}$ and NORMA$_{\mathrm{RI}}$. \textbf{d)} Performance Improvement with NORMA. Change in precision, sensitivity, specificity, and balanced accuracy (NORMA$_{\mathrm{RI}}$ versus Per$_{\mathrm{RI}}$); inner label shows the largest gain. \textbf{e)} Lead Time. Median lead time from NORMA$_{\mathrm{RI}}$ flag to first Pop$_{\mathrm{RI}}$ flag, by analyte (hours). Error bars denote the interquartile range.}
\label{fig:figure5}
\vspace{2pt}
\end{figure}
\clearpage

\begin{figure}[H]
\centering
\includegraphics[width=\textwidth,height=0.78\textheight,keepaspectratio]{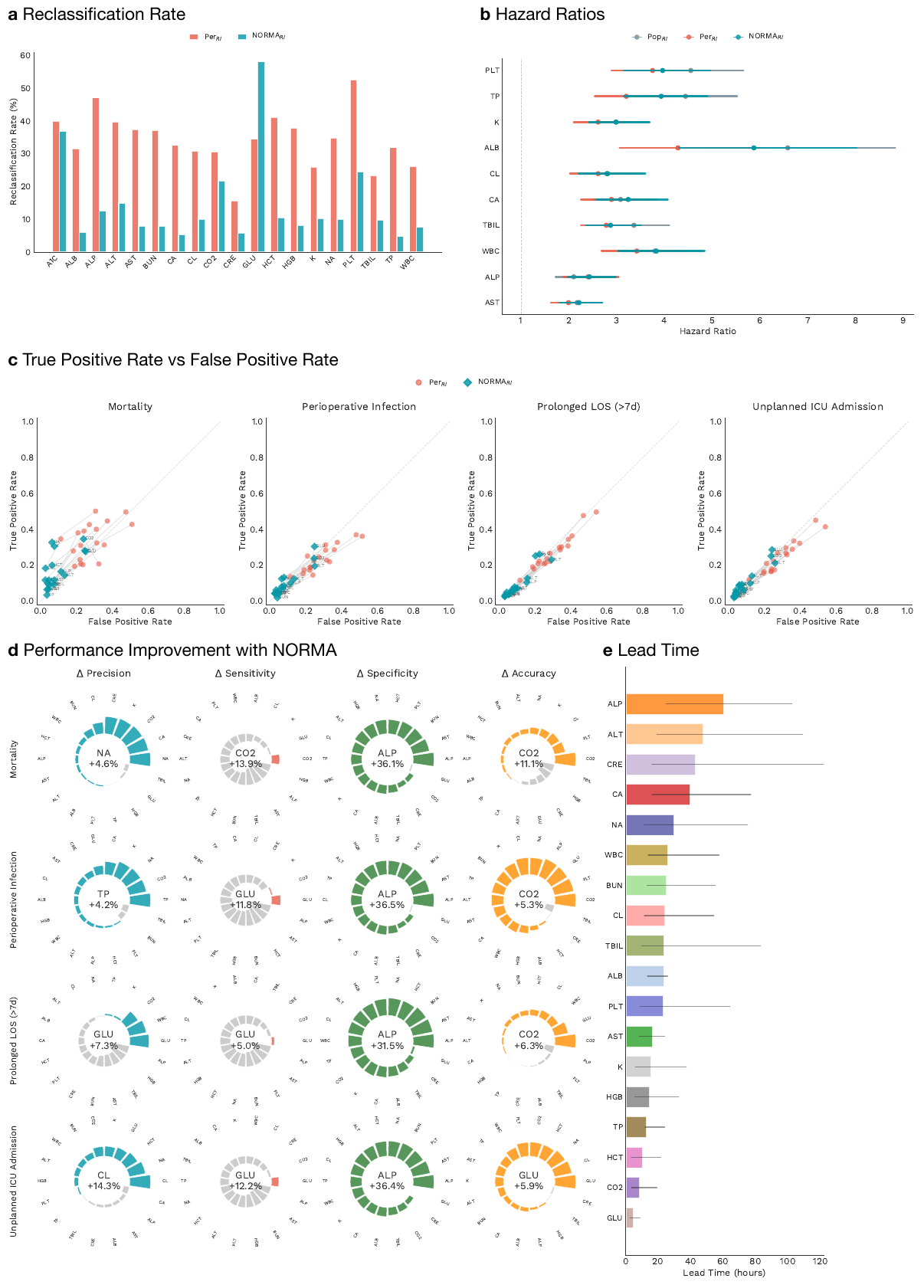}
\caption{\textbf{Reclassification performance and early risk stratification in the INSPIRE cohort.} \textbf{a)} Reclassification Rate. Reclassification rate by analyte among Pop$_{\mathrm{RI}}$-normal tests. \textbf{b)} Hazard Ratios. Cox hazard ratios for in-hospital mortality (adjusted for age and sex); analytes with $p < 0.05$ shown. \textbf{c)} True Positive Rate vs False Positive Rate. True positive rate versus false positive rate for clinical outcome prediction across in-hospital mortality, perioperative infection, prolonged hospital stay, and unplanned ICU admission, comparing Per$_{\mathrm{RI}}$ and NORMA$_{\mathrm{RI}}$. \textbf{d)} Performance Improvement with NORMA. Change in precision, sensitivity, specificity, and balanced accuracy (NORMA$_{\mathrm{RI}}$ versus Per$_{\mathrm{RI}}$); inner label shows the largest gain. \textbf{e)} Lead Time. Median lead time from NORMA$_{\mathrm{RI}}$ flag to first Pop$_{\mathrm{RI}}$ flag, by analyte (hours). Error bars denote the interquartile range.}
\label{fig:figure6}
\vspace{2pt}
\end{figure}
\clearpage

\section*{Supplementary Figures}
\setcounter{figure}{0}
\renewcommand{\figurename}{Supplementary Fig.}
\renewcommand{\thefigure}{\arabic{figure}}

\begin{figure}[H]
\centering
\includegraphics[width=\textwidth,height=0.78\textheight,keepaspectratio]{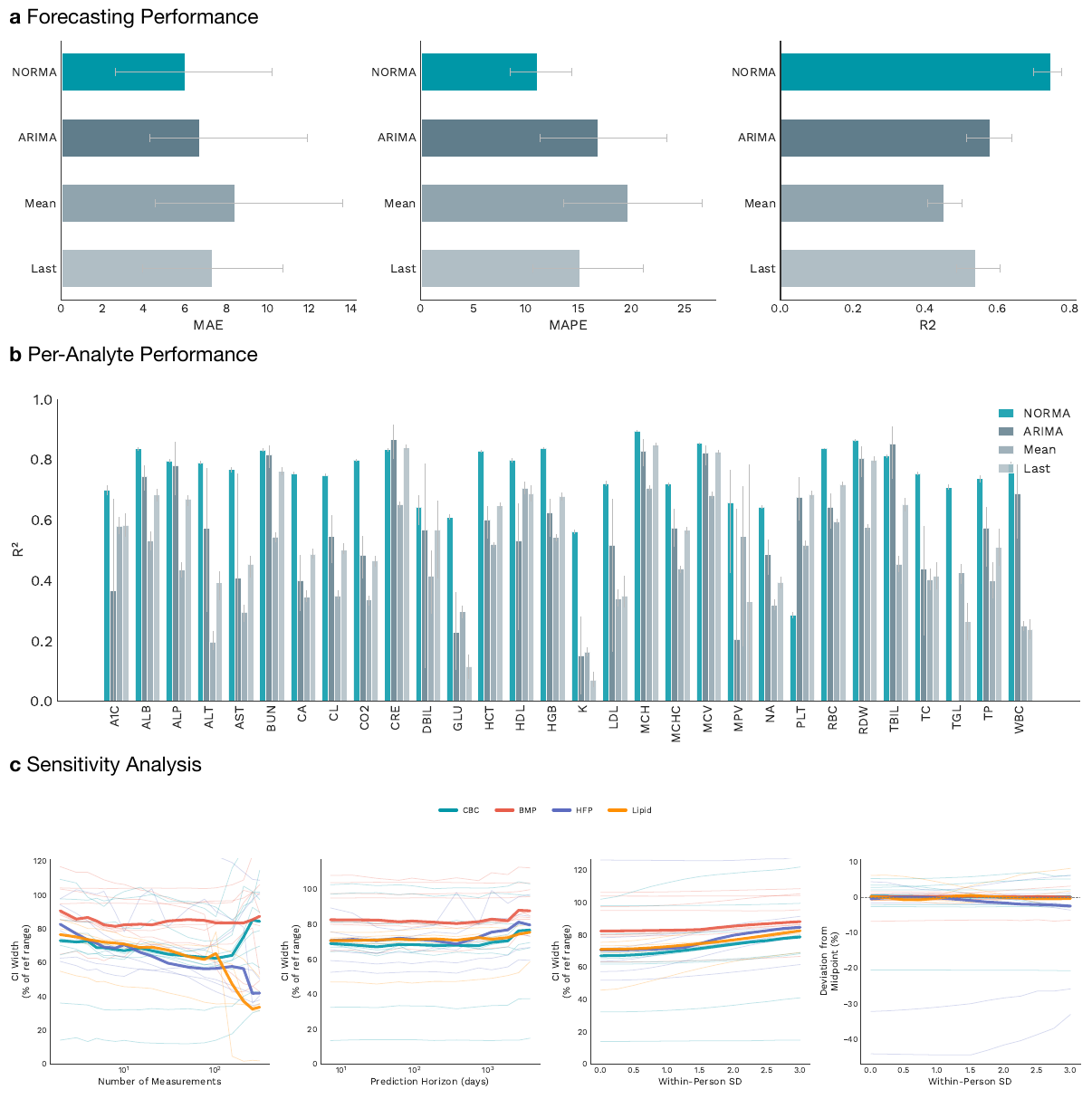}
\caption{\textbf{NORMA forecasting performance with Gaussian loss.} \textbf{a)} Forecasting Performance. Held-out test-set accuracy (mean absolute error, mean absolute percentage error, and coefficient of determination) for the Gaussian variant versus ARIMA, population mean, and last-value-carried-forward baselines. Error bars denote 95\% bootstrap confidence intervals. \textbf{b)} Per-Analyte Performance. Per-analyte coefficient of determination grouped by clinical panel. \textbf{c)} Sensitivity Analysis. Sensitivity of Gaussian-head prediction interval width and midpoint deviation (each as a percentage of Pop$_{\mathrm{RI}}$ width) to history length, forecast horizon, and within-person variability.}
\label{fig:supp_gaussian}
\vspace{2pt}
\end{figure}
\clearpage

\begin{figure}[H]
\centering
\includegraphics[width=\textwidth,height=0.78\textheight,keepaspectratio]{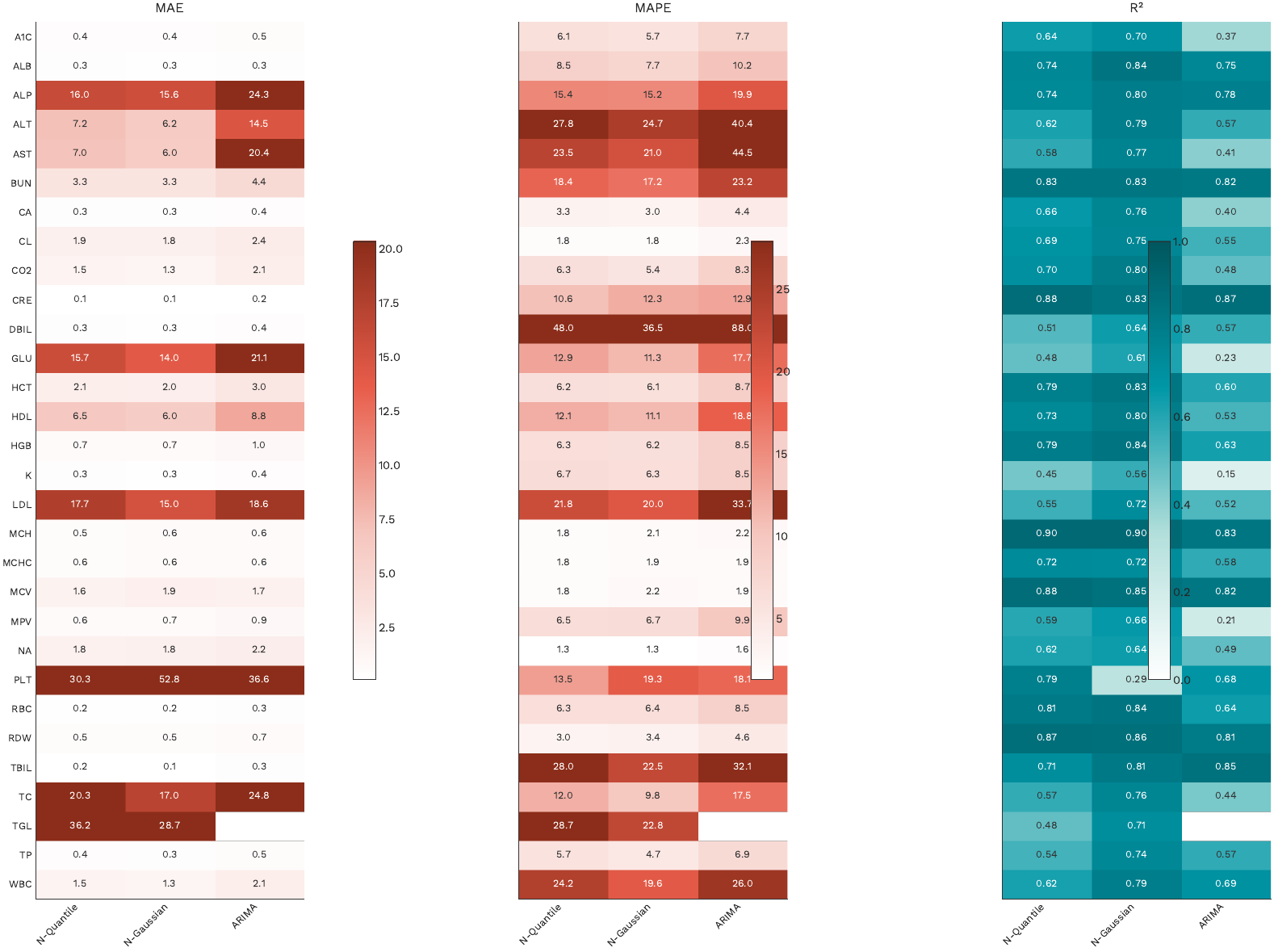}
\caption{\textbf{Per-analyte forecasting performance.} NORMA using quantile and Gaussian loss, compared to ARIMA, the best performing baseline. Color intensity encodes mean absolute error (MAE), mean absolute percentage error (MAPE), and coefficient of determination (R$^2$) on the held-out test set.}
\label{fig:analyte_heatmap}
\vspace{2pt}
\end{figure}
\clearpage

\begin{figure}[H]
\centering
\includegraphics[width=\textwidth,height=0.78\textheight,keepaspectratio]{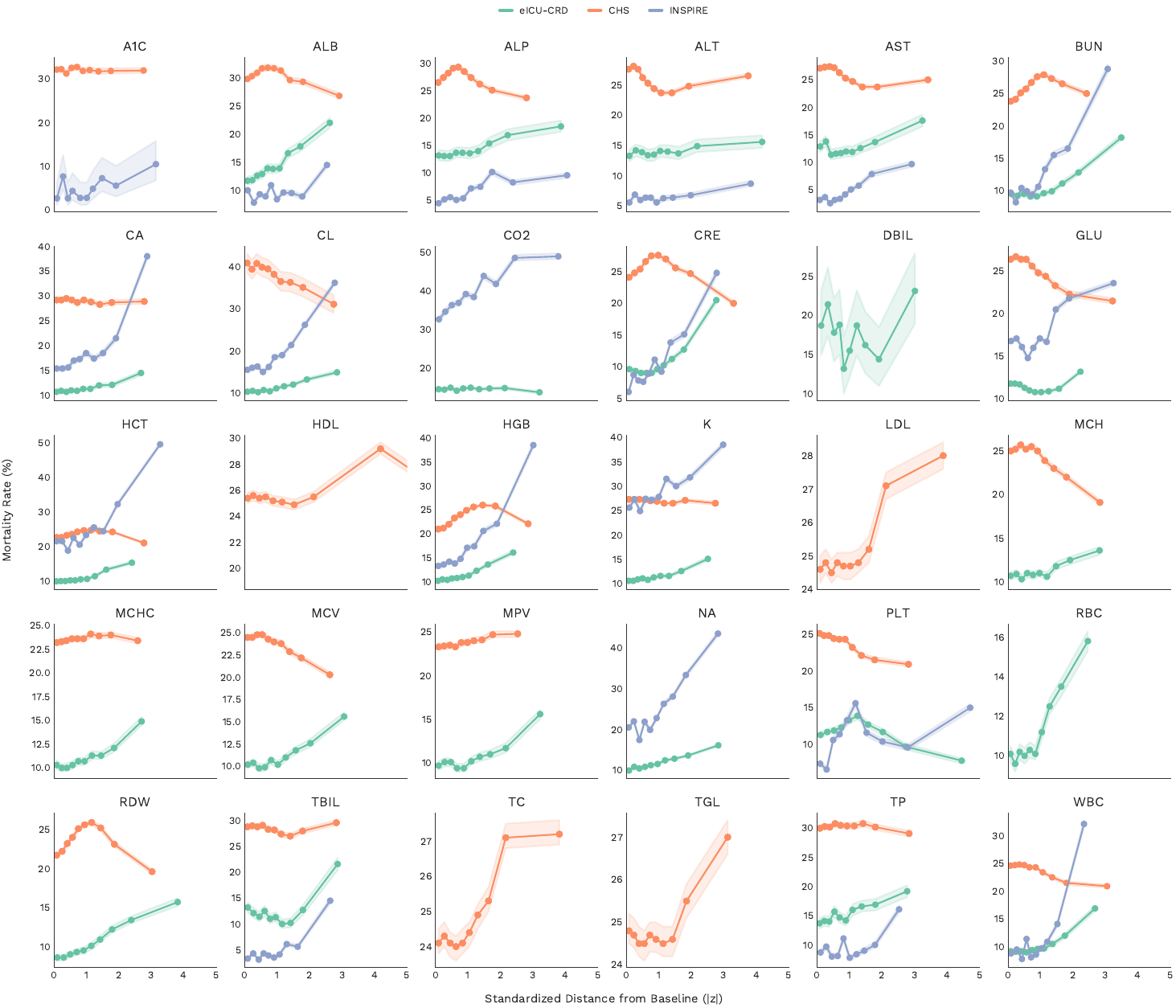}
\caption{\textbf{Association of absolute deviation from the personalized baseline with observed mortality.} Each panel shows the observed mortality rate by standardized absolute deviation from the personalized baseline, binned by decile, for Clalit Health Services, the eICU Collaborative Research Database, and the INSPIRE cohort.}
\label{fig:mortality_deviation}
\vspace{2pt}
\end{figure}
\clearpage

\begin{figure}[H]
\centering
\includegraphics[width=\textwidth,height=0.78\textheight,keepaspectratio]{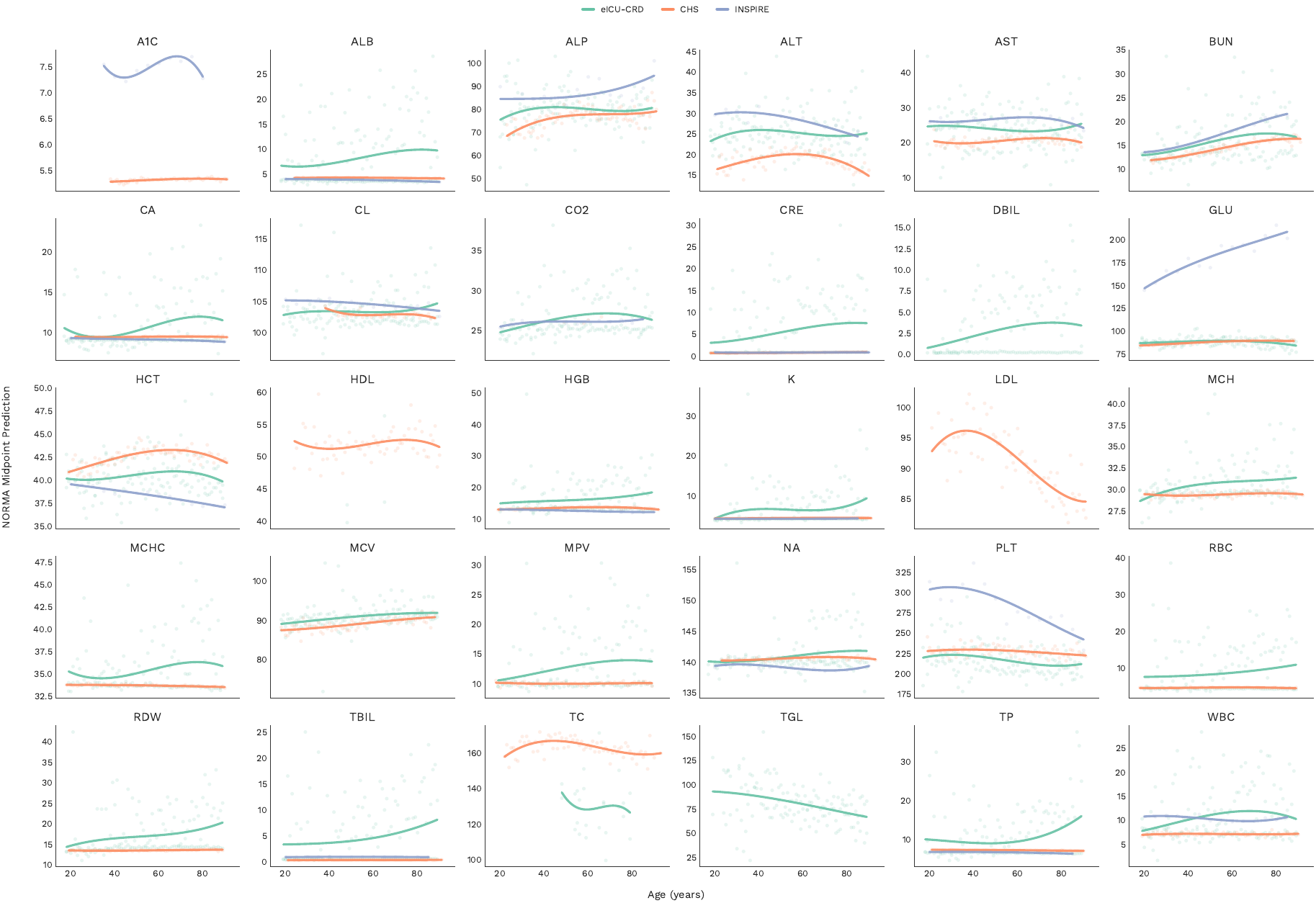}
\caption{\textbf{Age-dependent NORMA$_{\mathrm{RI}}$ midpoints.} Each panel shows individual midpoint predictions with a polynomial smooth overlaid, for Clalit Health Services, the eICU Collaborative Research Database, and the INSPIRE cohort.}
\label{fig:age_ri}
\vspace{2pt}
\end{figure}
\clearpage

\begin{figure}[H]
\centering
\includegraphics[width=\textwidth,height=0.78\textheight,keepaspectratio]{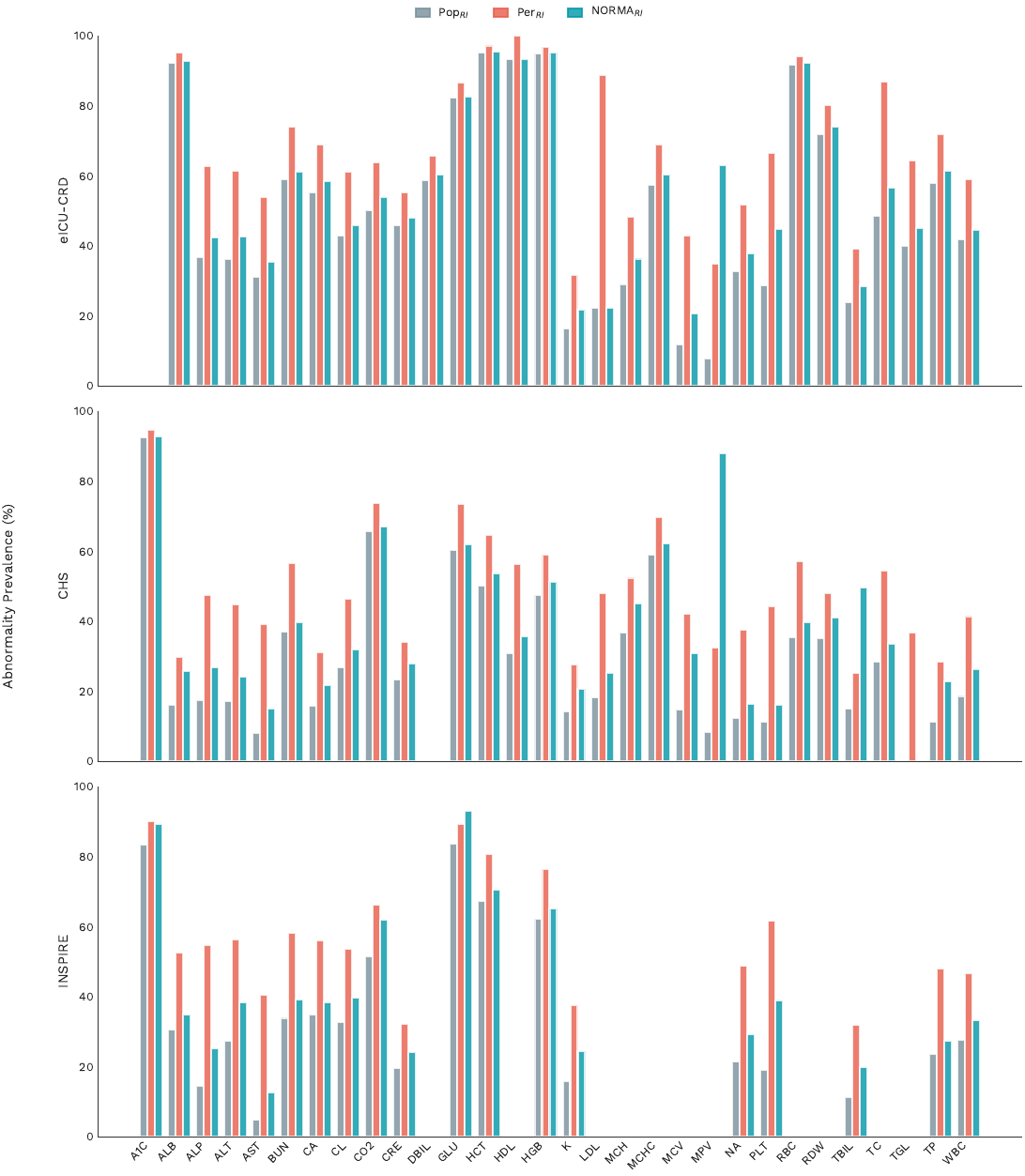}
\caption{\textbf{Abnormality prevalence by analyte and reference interval method.} Grouped bars show the percentage of measurements classified as abnormal by Pop$_{\mathrm{RI}}$, Per$_{\mathrm{RI}}$, and NORMA$_{\mathrm{RI}}$, for the eICU Collaborative Research Database, Clalit Health Services, and the INSPIRE cohort.}
\label{fig:prevalence_by_analyte}
\vspace{2pt}
\end{figure}
\clearpage

\begin{figure}[H]
\centering
\includegraphics[width=\textwidth,height=0.78\textheight,keepaspectratio]{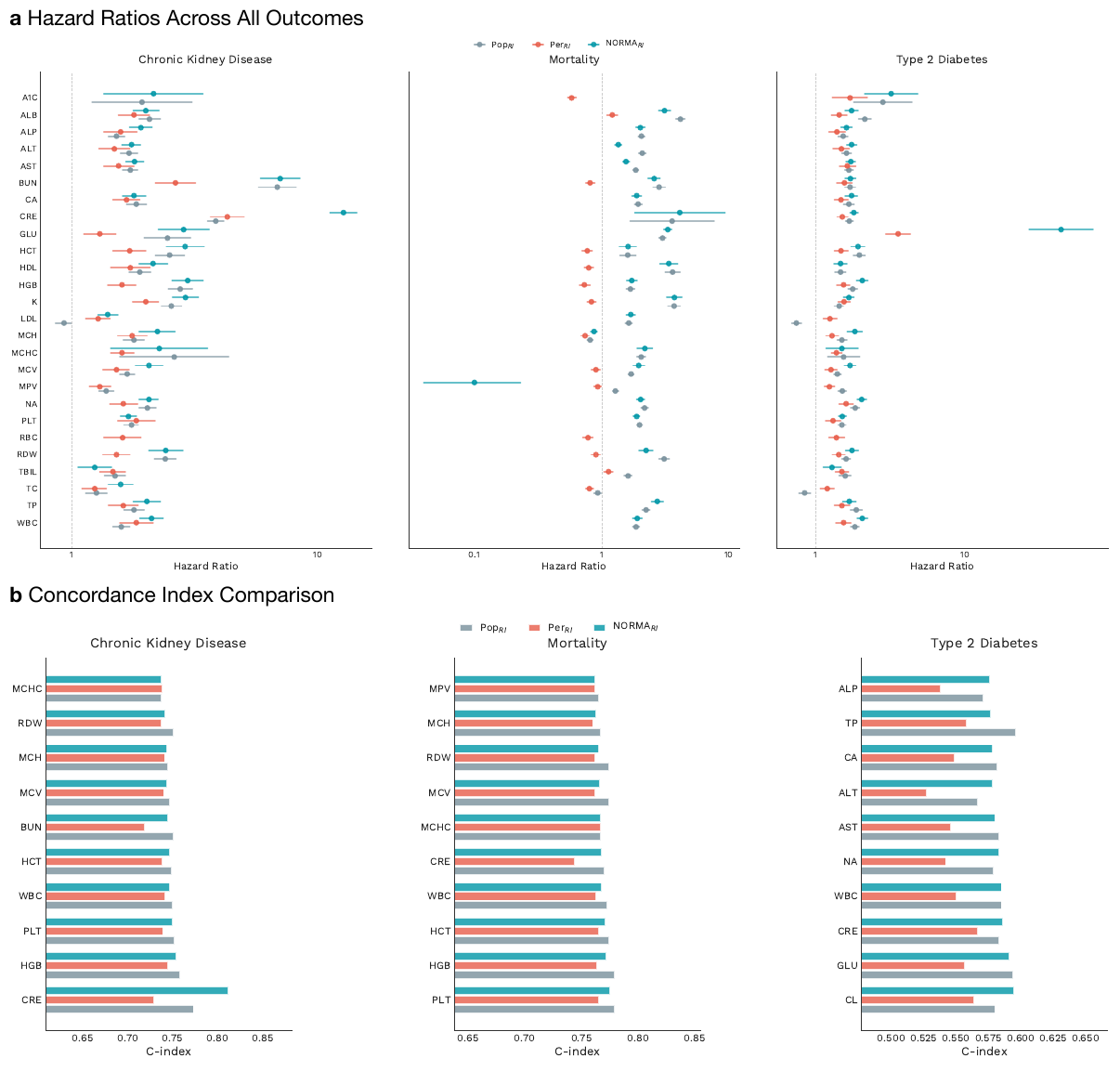}
\caption{\textbf{Proportional hazards analysis in Clalit Health Services.} \textbf{a)} Hazard Ratios Across All Outcomes. Cox hazard ratios for chronic kidney disease, all-cause mortality, type 2 diabetes, and anemia, comparing Pop$_{\mathrm{RI}}$, Per$_{\mathrm{RI}}$, and NORMA$_{\mathrm{RI}}$ abnormality flags; only analytes with $p < 0.05$ are shown. \textbf{b)} Concordance Index Comparison. Concordance index for the top 10 analytes ranked by NORMA$_{\mathrm{RI}}$ concordance per outcome.}
\label{fig:supp_cox_chs}
\vspace{2pt}
\end{figure}
\clearpage

\begin{figure}[H]
\centering
\includegraphics[width=\textwidth,height=0.78\textheight,keepaspectratio]{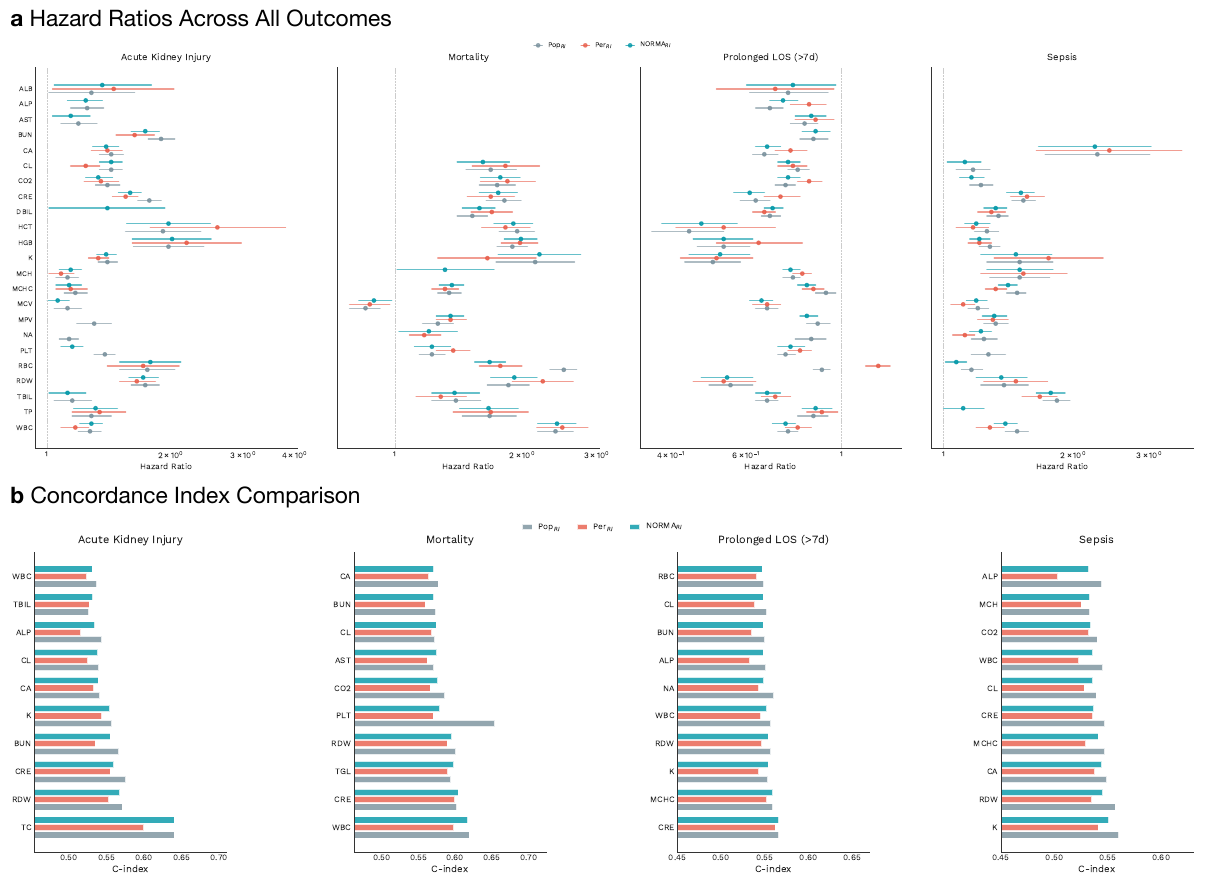}
\caption{\textbf{Proportional hazards analysis in the eICU Collaborative Research Database.} \textbf{a)} Hazard Ratios Across All Outcomes. Cox hazard ratios for acute kidney injury, in-hospital mortality, prolonged ICU stay, and sepsis, comparing Pop$_{\mathrm{RI}}$, Per$_{\mathrm{RI}}$, and NORMA$_{\mathrm{RI}}$ abnormality flags; only analytes with $p < 0.05$ are shown. \textbf{b)} Concordance Index Comparison. Concordance index for the top 10 analytes ranked by NORMA$_{\mathrm{RI}}$ concordance per outcome.}
\label{fig:supp_cox_eicu}
\vspace{2pt}
\end{figure}
\clearpage

\begin{figure}[H]
\centering
\includegraphics[width=\textwidth,height=0.78\textheight,keepaspectratio]{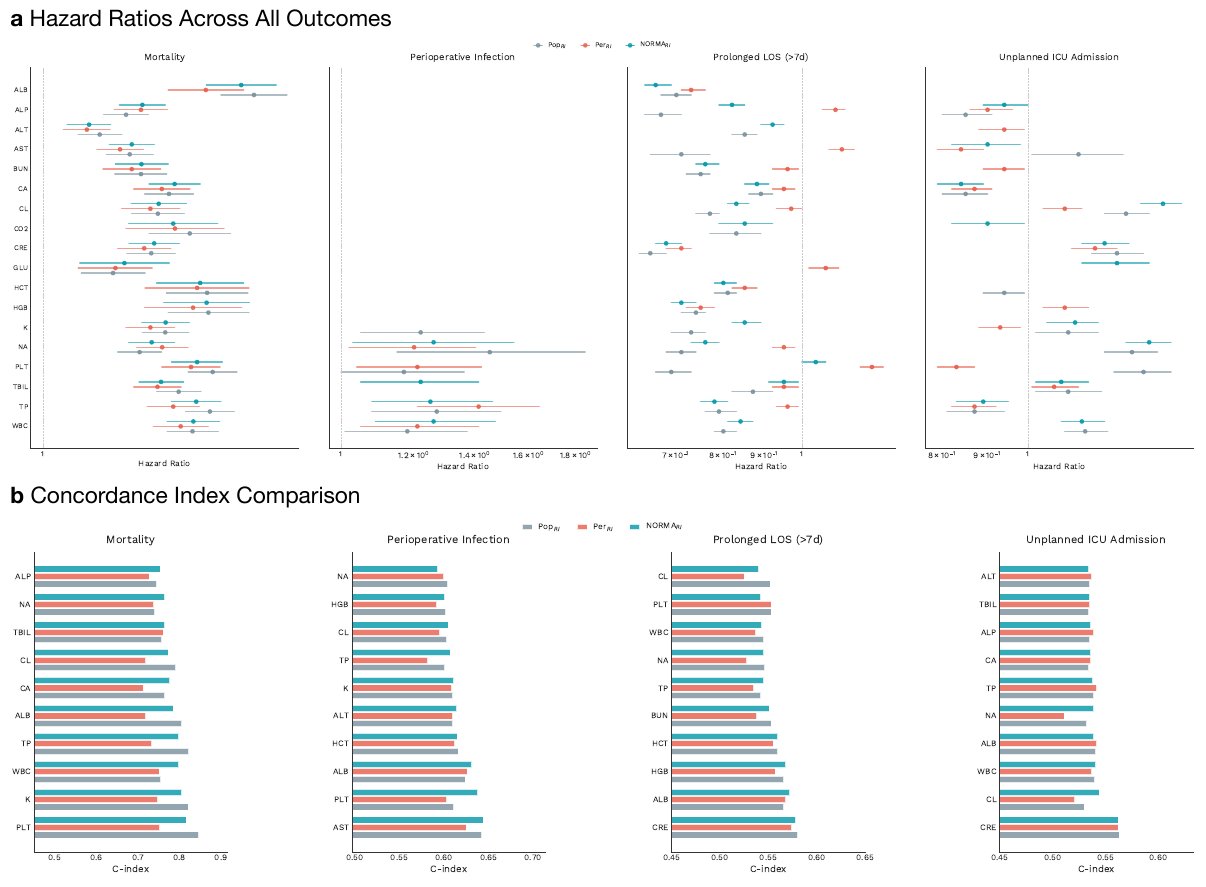}
\caption{\textbf{Proportional hazards analysis in the INSPIRE cohort.} \textbf{a)} Hazard Ratios Across All Outcomes. Cox hazard ratios for in-hospital mortality, perioperative infection, prolonged hospital stay, and unplanned ICU admission, comparing Pop$_{\mathrm{RI}}$, Per$_{\mathrm{RI}}$, and NORMA$_{\mathrm{RI}}$ abnormality flags; only analytes with $p < 0.05$ are shown. \textbf{b)} Concordance Index Comparison. Concordance index for the top 10 analytes ranked by NORMA$_{\mathrm{RI}}$ concordance per outcome.}
\label{fig:supp_cox_inspire}
\vspace{2pt}
\end{figure}
\clearpage

\begin{figure}[H]
\centering
\includegraphics[width=\textwidth,height=0.78\textheight,keepaspectratio]{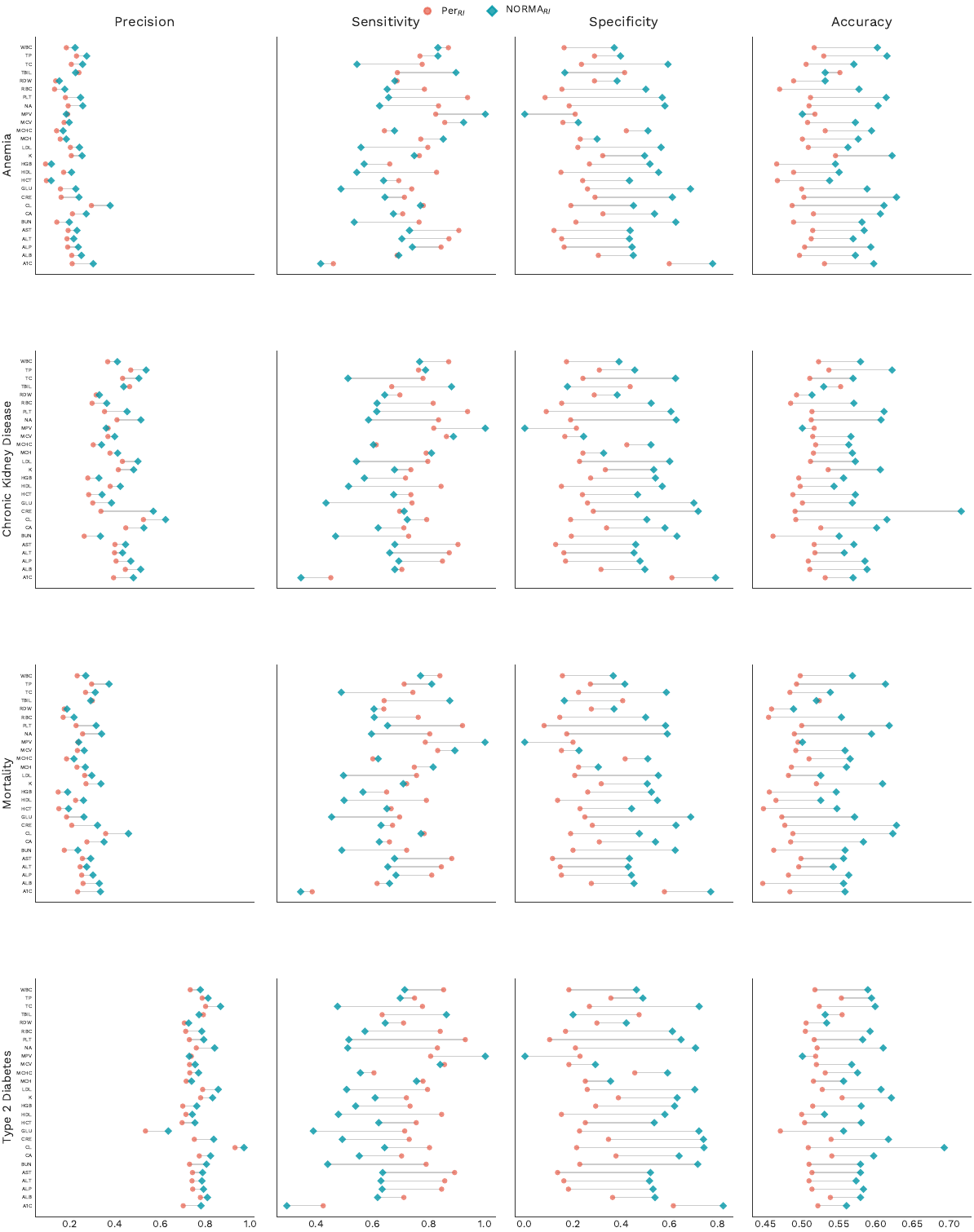}
\caption{\textbf{Clinical outcome prediction performance in Clalit Health Services.} Each point shows Per$_{\mathrm{RI}}$ and NORMA$_{\mathrm{RI}}$ values for precision, sensitivity, specificity, and balanced accuracy across outcomes. Connected pairs show the direction and magnitude of change between methods for each analyte. Analytes with fewer than 100 measurements are omitted.}
\label{fig:eval_dotplot_chs}
\vspace{2pt}
\end{figure}
\clearpage

\begin{figure}[H]
\centering
\includegraphics[width=\textwidth,height=0.78\textheight,keepaspectratio]{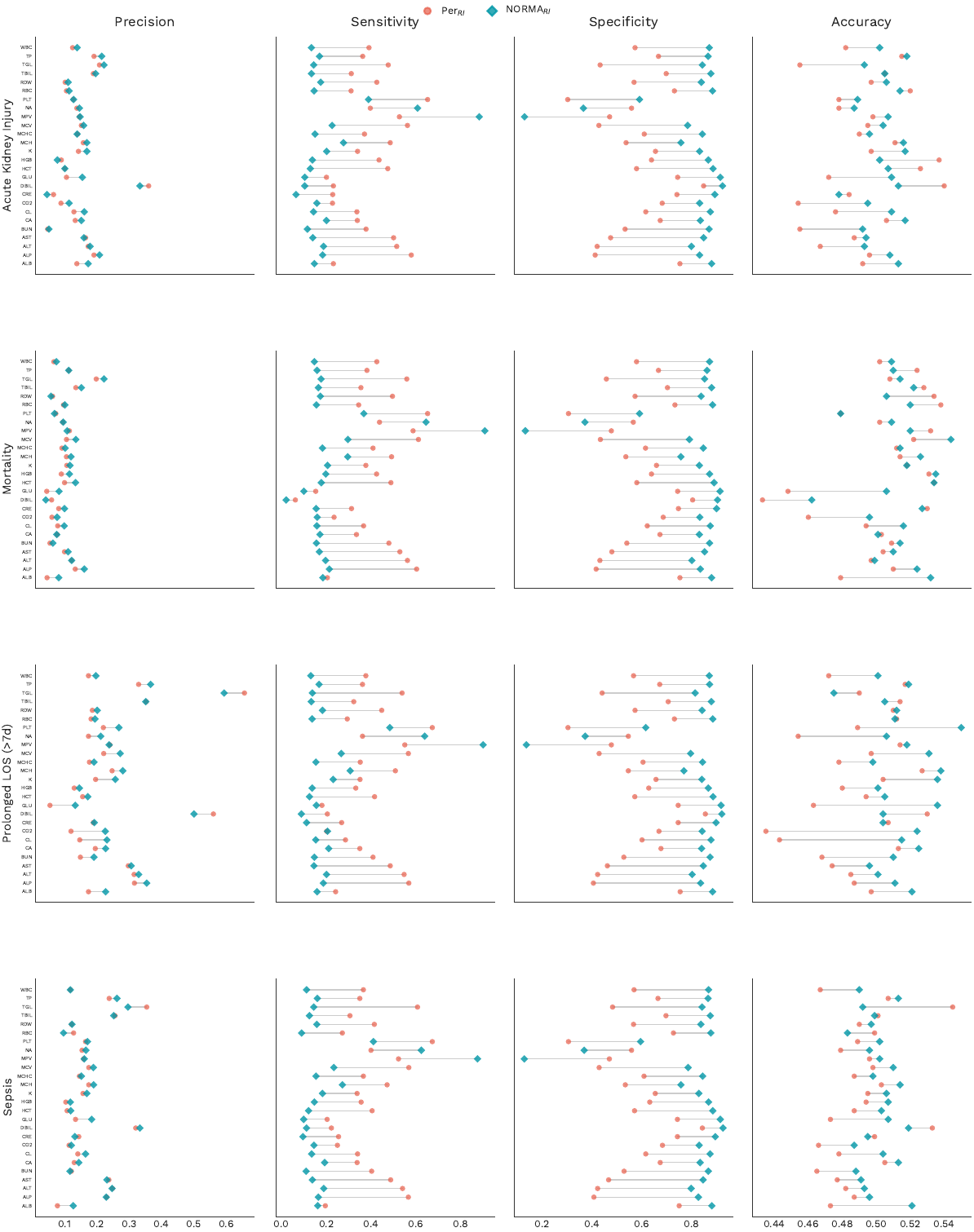}
\caption{\textbf{Clinical outcome prediction performance in the eICU Collaborative Research Database.} Each point shows Per$_{\mathrm{RI}}$ and NORMA$_{\mathrm{RI}}$ values for precision, sensitivity, specificity, and balanced accuracy across outcomes. Connected pairs show the direction and magnitude of change between methods for each analyte. Analytes with fewer than 100 measurements are omitted.}
\label{fig:eval_dotplot_eicu}
\vspace{2pt}
\end{figure}
\clearpage

\begin{figure}[H]
\centering
\includegraphics[width=\textwidth,height=0.78\textheight,keepaspectratio]{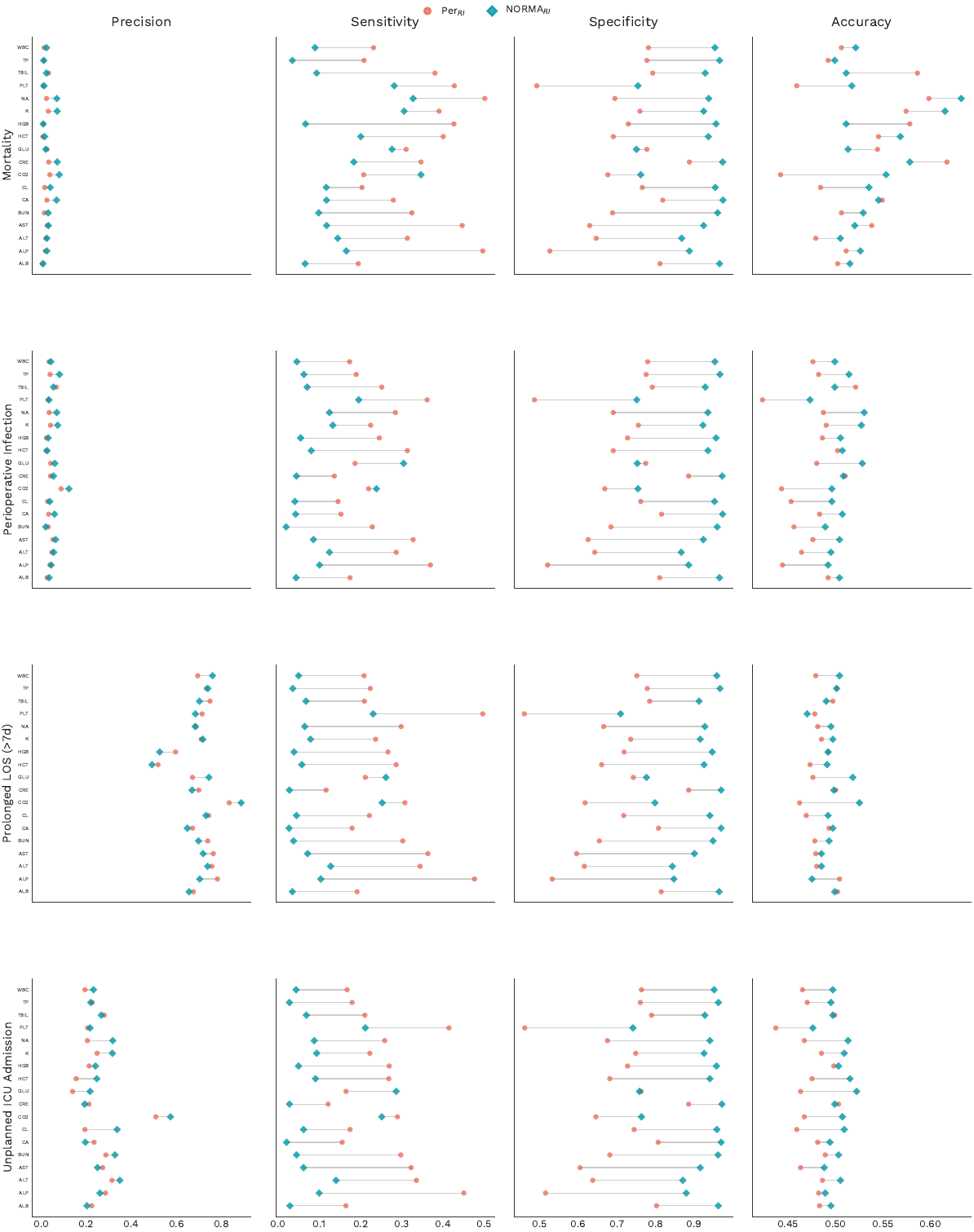}
\caption{\textbf{Clinical outcome prediction performance in the INSPIRE cohort.} Each point shows Per$_{\mathrm{RI}}$ and NORMA$_{\mathrm{RI}}$ values for precision, sensitivity, specificity, and balanced accuracy across outcomes. Connected pairs show the direction and magnitude of change between methods for each analyte. Analytes with fewer than 100 measurements are omitted.}
\label{fig:eval_dotplot_inspire}
\vspace{2pt}
\end{figure}
\clearpage

\section*{Supplementary Tables}
\setcounter{table}{0}
\renewcommand{\tablename}{Supplementary Table}
\renewcommand{\thetable}{\arabic{table}}

\begin{table}[ht]
\centering
\caption{\textbf{Analyte reference information.} Full analyte name, abbreviation, unit of measurement, and conventional Pop$_{\mathrm{RI}}$ for all 30 laboratory analytes included in the study. Pop$_{\mathrm{RI}}$ were obtained from the American Board of Internal Medicine Laboratory Test Reference Ranges (January 2025).}
\adjustbox{max width=\textwidth}{%
\small
\begin{tabular}{llll}
\toprule
Analyte & Full Name & Unit & Population RI \\
\midrule
A1C & Hemoglobin A1c & \% & 4.0--5.6 \\
ALB & Albumin & g/dL & 3.5--5.5 \\
ALP & Alkaline Phosphatase & U/L & 30--120 \\
ALT & Alanine Aminotransferase & U/L & 0--35 \\
AST & Aspartate Aminotransferase & U/L & 0--35 \\
BUN & Blood Urea Nitrogen & mg/dL & 8--20 \\
CA & Calcium & mg/dL & 8.6--10.2 \\
CL & Chloride & mEq/L & 98--106 \\
CO2 & Bicarbonate & mEq/L & 23--28 \\
CRE & Creatinine & mg/dL & F: 0.5--1.1, M: 0.7--1.3 \\
DBIL & Direct Bilirubin & mg/dL & 0.1--0.3 \\
GLU & Glucose & mg/dL & 70--99 \\
HCT & Hematocrit & \% & F: 37--47, M: 42--50 \\
HDL & HDL Cholesterol & mg/dL & 40--100 \\
HGB & Hemoglobin & g/dL & F: 12.0--16.0, M: 14.0--18.0 \\
K & Potassium & mEq/L & 3.5--5.0 \\
LDL & LDL Cholesterol & mg/dL & <130 \\
MCH & Mean Corpuscular Hemoglobin & pg & 28--32 \\
MCHC & MCH Concentration & g/dL & 33--36 \\
MCV & Mean Corpuscular Volume & fL & 80--98 \\
MPV & Mean Platelet Volume & fL & 7.5--12.5 \\
NA & Sodium & mEq/L & 136--145 \\
PLT & Platelet Count & 10$^3$/$\mu$L & 150--450 \\
RBC & Red Blood Cell Count & 10$^6$/$\mu$L & F: 4.0--5.2, M: 4.5--5.9 \\
RDW & Red Cell Distribution Width & \% & 9--14.5 \\
TBIL & Total Bilirubin & mg/dL & 0.3--1.0 \\
TC & Total Cholesterol & mg/dL & 100--200 \\
TGL & Triglycerides & mg/dL & <150 \\
TP & Total Protein & g/dL & 6.0--8.0 \\
WBC & White Blood Cell Count & 10$^3$/$\mu$L & 4.5--11.0 \\
\bottomrule
\end{tabular}%
}
\label{tab:analyte_reference}
\end{table}
\clearpage

\begin{table}[ht]
\centering
\caption{\textbf{Cohort characteristics.} Patient counts, demographics (age, sex, follow-up duration), and per-analyte measurement counts and mean values (± standard deviation) across all five cohorts: EHRSHOT, MIMIC-IV, Clalit Health Services, the eICU Collaborative Research Database, and the INSPIRE cohort.}
\adjustbox{max width=\textwidth}{%
\small
\begin{tabular}{lrrrrrrrrrr}
\toprule
  & \multicolumn{2}{c}{EHRSHOT} & \multicolumn{2}{c}{MIMIC-IV} & \multicolumn{2}{c}{eICU-CRD} & \multicolumn{2}{c}{CHS} & \multicolumn{2}{c}{INSPIRE} \\
\cmidrule(lr){2-3} \cmidrule(lr){4-5} \cmidrule(lr){6-7} \cmidrule(lr){8-9} \cmidrule(lr){10-11}
\textit{Patients} & \multicolumn{2}{c}{5,676} & \multicolumn{2}{c}{179,601} & \multicolumn{2}{c}{98,432} & \multicolumn{2}{c}{1,450,862} & \multicolumn{2}{c}{51,159} \\
\textit{Age} & \multicolumn{2}{c}{54.9 $\pm$ 17.2} & \multicolumn{2}{c}{58.2 $\pm$ 19.0} & \multicolumn{2}{c}{63.3 $\pm$ 16.1} & \multicolumn{2}{c}{55.1 $\pm$ 18.6} & \multicolumn{2}{c}{57.6 $\pm$ 14.9} \\
\textit{Sex} & \multicolumn{2}{c}{52\% F / 48\% M} & \multicolumn{2}{c}{54\% F / 46\% M} & \multicolumn{2}{c}{46\% F / 54\% M} & \multicolumn{2}{c}{63\% F / 37\% M} & \multicolumn{2}{c}{52\% F / 48\% M} \\
\textit{Time} & \multicolumn{2}{c}{2.49 [0.28, 7.20] yr} & \multicolumn{2}{c}{1.01 [0.06, 4.08] yr} & \multicolumn{2}{c}{0.02 [0.01, 0.03] yr} & \multicolumn{2}{c}{20.24 [17.51, 21.47] yr} & \multicolumn{2}{c}{0.48 [0.17, 0.84] yr} \\
\midrule
Analyte & N & Value & N & Value & N & Value & N & Value & N & Value \\
\midrule
A1C & 1,745 & 6.3 $\pm$ 1.2 & 39,996 & 6.6 $\pm$ 1.3 & --- & --- & 336,493 & 7.1 $\pm$ 1.3 & 391 & 6.7 $\pm$ 1.0 \\
ALB & 4,282 & 3.3 $\pm$ 0.8 & 55,220 & 3.8 $\pm$ 0.8 & 19,075 & 2.6 $\pm$ 0.7 & 790,253 & 4.1 $\pm$ 0.5 & 30,728 & 3.4 $\pm$ 0.6 \\
\bottomrule
\end{tabular}%
}
\label{tab:cohort}
\end{table}
\clearpage

\begin{table}[H]
    \centering
\caption{\textbf{NORMA architectural and training design choices.} Summary of configurable components including temporal encoding, health state encoding, age and laboratory value representations, output parameterization, and context token usage.}
\label{tab:norma_design}
    \adjustbox{max width=\textwidth}{%
\small
\begin{tabular}{p{2.8cm} p{2.8cm} p{8.0cm}}
    \toprule
    \textbf{Component} & \textbf{Design Choice} & \textbf{Description} \\
    \midrule
    \multirow{2}{*}{Time embedding}
    & Time2Vec
    & Encodes the number of days elapsed since the first measurement using learned linear and sinusoidal functions. Captures long-term temporal trends and potential periodic patterns. \\
    \addlinespace
    & Delta-$t$
    & Encodes the log-compressed time difference between consecutive measurements. Explicitly represents irregular sampling and emphasizes how much time has passed since the last value. \\
    \midrule
    \multirow{2}{*}{Health state encoding}
    & Binary
    & Encodes each lab value as either within or outside the population reference range; does not distinguish high vs.\ low and can struggle with J-shaped biomarkers. \\
    \addlinespace
    & Ternary
    & Encodes whether the lab value is low, normal, or high relative to the reference range; handles asymmetric risk patterns; however, can increase overfitting risk. \\
    \midrule
    \multirow{2}{*}{Age encoding}
    & Raw
    & Uses age directly as a numeric input; assumes linear effects. \\
    \addlinespace
    & Binned
    & Discretizes age into groups (e.g., decades) and maps each bin to a learned embedding vector; captures nonlinear and threshold effects. \\
    \midrule
    \multirow{2}{*}{Lab value encoding}
    & Raw
    & Uses observed biomarker values directly as model input; preserves absolute clinical scale and keeps preprocessing minimal. \\
\bottomrule
    \end{tabular}%
}
    \end{table}
    
\clearpage

\begin{table}[ht]
\centering
\caption{\textbf{Within-person and between-person biological variability.} Per-analyte within-person coefficient of variation, between-person coefficient of variation, and individuality index for the eICU Collaborative Research Database and Clalit Health Services. Values are reported as median [95\% confidence interval].}
\adjustbox{max width=\textwidth}{%
\small
\begin{tabular}{lrrrrrrrrr}
\toprule
  & \multicolumn{3}{c}{eICU-CRD} & \multicolumn{3}{c}{CHS} & \multicolumn{3}{c}{INSPIRE} \\
\cmidrule(lr){2-4} \cmidrule(lr){5-7} \cmidrule(lr){8-10}
Analyte & CV$_{\text{intra}}$ & CV$_{\text{inter}}$ & II & CV$_{\text{intra}}$ & CV$_{\text{inter}}$ & II & CV$_{\text{intra}}$ & CV$_{\text{inter}}$ & II \\
\midrule
A1C & --- & --- & --- & 8.6 [8.5, 8.6] & 16.5 [16.5, 16.6] & 0.5 [0.5, 0.5] & 4.8 [4.4, 5.3] & 13.0 [11.8, 13.8] & 0.4 [0.3, 0.4] \\
ALB & 8.9 [8.8, 8.9] & 22.6 [22.4, 22.7] & 0.4 [0.4, 0.4] & 6.7 [6.7, 6.8] & 18.9 [18.3, 19.4] & 0.4 [0.3, 0.4] & 8.4 [8.3, 8.4] & 13.4 [13.3, 13.5] & 0.6 [0.6, 0.6] \\
ALP & 10.6 [10.5, 10.7] & 42.1 [41.8, 42.4] & 0.2 [0.2, 0.3] & 17.8 [17.7, 17.8] & 30.1 [30.0, 30.2] & 0.6 [0.6, 0.6] & 10.9 [10.7, 11.0] & 34.0 [33.7, 34.4] & 0.3 [0.3, 0.3] \\
ALT & 15.5 [15.3, 15.7] & 65.3 [64.8, 65.8] & 0.2 [0.2, 0.2] & 34.2 [34.2, 34.3] & 50.2 [50.0, 50.5] & 0.7 [0.7, 0.7] & 19.2 [19.0, 19.4] & 49.8 [49.3, 50.3] & 0.4 [0.4, 0.4] \\
AST & 18.3 [18.1, 18.5] & 63.4 [62.9, 63.8] & 0.3 [0.3, 0.3] & 24.7 [24.6, 24.7] & 35.1 [34.9, 35.4] & 0.7 [0.7, 0.7] & 14.4 [14.3, 14.6] & 32.6 [32.2, 32.9] & 0.4 [0.4, 0.5] \\
BUN & 16.5 [16.4, 16.6] & 58.7 [58.5, 58.9] & 0.3 [0.3, 0.3] & 20.5 [20.4, 20.5] & 36.2 [36.2, 36.3] & 0.6 [0.6, 0.6] & 15.4 [15.2, 15.5] & 42.0 [41.5, 42.4] & 0.4 [0.4, 0.4] \\
\bottomrule
\end{tabular}%
}
\label{tab:variability}
\end{table}
\clearpage

\begin{table}[ht]
\centering
\caption{\textbf{NORMA forecasting performance.} Mean absolute error (MAE), mean absolute percentage error (MAPE), and coefficient of determination (R$^2$) for NORMA Gaussian, NORMA quantile, ARIMA, population mean, and last-value-carried-forward, reported on training, validation, and held-out test sets. Values are reported as median [interquartile range] across analytes.}
\adjustbox{max width=\textwidth}{%
\small
\begin{tabular}{llrrrrr}
\toprule
Metric & & NORMA Gaussian & NORMA Quantile & ARIMA & Mean & Last \\
\midrule
\multirow{3}{*}{MAE} & Train & 6.0 [3.2, 11.1] & \textbf{5.7 [2.9, 10.1]} & 6.8 [2.5, 10.2] & 8.4 [4.5, 15.0] & 7.4 [3.7, 11.0] \\
 & Val & 6.0 [2.6, 10.7] & \textbf{5.8 [2.9, 8.7]} & 6.8 [3.6, 10.0] & 8.3 [4.9, 13.5] & 7.3 [3.8, 10.3] \\
 & Test & 6.0 [2.6, 10.2] & \textbf{5.9 [3.6, 9.1]} & 6.7 [4.3, 11.9] & 8.4 [4.5, 13.6] & 7.3 [3.9, 10.7] \\
\midrule
\multirow{3}{*}{MAPE} & Train & \textbf{10.9 [8.8, 14.0]} & 11.9 [8.6, 14.7] & 15.9 [11.3, 20.1] & 19.4 [14.8, 25.4] & 14.8 [10.9, 19.6] \\
 & Val & \textbf{11.1 [7.9, 13.2]} & 12.6 [8.7, 15.6] & 15.7 [12.1, 21.5] & 19.7 [14.0, 27.4] & 15.2 [10.6, 20.1] \\
 & Test & \textbf{11.1 [8.5, 14.3]} & 12.3 [7.9, 17.3] & 16.9 [11.3, 23.3] & 19.7 [13.5, 26.7] & 15.2 [10.6, 21.1] \\
\midrule
\multirow{3}{*}{R2} & Train & \textbf{0.76 [0.72, 0.80]} & 0.70 [0.66, 0.75] & 0.62 [0.57, 0.70] & 0.46 [0.41, 0.49] & 0.56 [0.47, 0.62] \\
 & Val & \textbf{0.75 [0.69, 0.78]} & 0.69 [0.66, 0.73] & 0.62 [0.53, 0.67] & 0.46 [0.43, 0.49] & 0.55 [0.49, 0.62] \\
 & Test & \textbf{0.75 [0.70, 0.78]} & 0.68 [0.64, 0.72] & 0.58 [0.51, 0.64] & 0.45 [0.41, 0.50] & 0.54 [0.49, 0.61] \\
\bottomrule
\end{tabular}%
}
\label{tab:prediction_performance}
\end{table}
\clearpage

\begin{table}[ht]
\centering
\caption{\textbf{Sensitivity of NORMA prediction intervals to patient features.} Median and interquartile range of prediction interval width change (as a percentage of Pop$_{\mathrm{RI}}$ width) and midpoint shift (as a percentage of Pop$_{\mathrm{RI}}$ width) in response to history length, forecast horizon, and within-person variability, for both the quantile and Gaussian parameterizations.}
\adjustbox{max width=\textwidth}{%
\small
\begin{tabular}{llrrrr}
\toprule
 & & \multicolumn{2}{c}{CI Width Change (\%)} & \multicolumn{2}{c}{Midpoint Shift (\%)} \\
\cmidrule(lr){3-4} \cmidrule(lr){5-6}
Model & Feature & Median & IQR & Median & IQR \\
\midrule
Quantile & History Length & 4.8 & [1.0, 14.4] & 0.0 & [0.0, 0.0] \\
Quantile & Prediction Horizon & 1.6 & [0.1, 11.1] & 0.0 & [0.0, 0.0] \\
Quantile & Within-Person Variability & 115.2 & [111.4, 120.5] & 0.7 & [0.3, 1.7] \\
\midrule
Gaussian & History Length & 28.3 & [20.6, 41.8] & 12.8 & [5.6, 20.6] \\
Gaussian & Prediction Horizon & 7.2 & [3.0, 9.9] & 1.6 & [0.8, 3.2] \\
Gaussian & Within-Person Variability & 7.2 & [3.7, 12.4] & 1.3 & [0.3, 3.6] \\
\bottomrule
\end{tabular}%
}
\label{tab:sensitivity_summary}
\end{table}
\clearpage

\begin{table}[ht]
\centering
\caption{\textbf{Abnormality prevalence and reclassification rates.} Overall percentage of measurements classified as abnormal by Pop$_{\mathrm{RI}}$,  Per$_{\mathrm{RI}}$, and NORMA$_{\mathrm{RI}}$, and the reclassification rate (the proportion of Pop$_{\mathrm{RI}}$-normal tests flagged abnormal by Per$_{\mathrm{RI}}$ or NORMA$_{\mathrm{RI}}$) in the eICU Collaborative Research Database and Clalit Health Services.}
\adjustbox{max width=\textwidth}{%
\small
\begin{tabular}{lrrrrrr}
\toprule
Dataset & N & Pop$_{RI}$ (\%) & Per$_{RI}$ (\%) & NORMA$_{RI}$ (\%) & Per$_{RI}$ RR (\%) & NORMA$_{RI}$ RR (\%) \\
\midrule
eICU-CRD & 29 & 50.2 & 68.1 & 55.8 & 37.3 & 9.2 \\
CHS & 31 & 29.6 & 46.8 & 39.1 & 27.2 & 12.0 \\
INSPIRE & 19 & 35.1 & 57.0 & 42.5 & 34.3 & 14.4 \\
\bottomrule
\end{tabular}%
}
\label{tab:prevalence}
\end{table}
\clearpage

\begin{table}[ht]
\centering
\caption{\textbf{Per-analyte reclassification.} Among tests with a Per$_{\mathrm{RI}}$ setpoint within Pop$_{\mathrm{RI}}$, the number per 1,000 flagged abnormal by Pop$_{\mathrm{RI}}$, Per$_{\mathrm{RI}}$, and NORMA$_{\mathrm{RI}}$ for each analyte, in the eICU Collaborative Research Database and Clalit Health Services.}
\adjustbox{max width=\textwidth}{%
\small
\begin{tabular}{lrrrrrrrrr}
\toprule
 & \multicolumn{3}{c}{CHS} & \multicolumn{3}{c}{eICU-CRD} & \multicolumn{3}{c}{INSPIRE} \\
\cmidrule(lr){2-4} \cmidrule(lr){5-7} \cmidrule(lr){8-10}
Analyte & Pop$_{RI}$ & Per$_{RI}$ & NORMA$_{RI}$ & Pop$_{RI}$ & Per$_{RI}$ & NORMA$_{RI}$ & Pop$_{RI}$ & Per$_{RI}$ & NORMA$_{RI}$ \\
\midrule
A1C & 240 & 315 & 282 & --- & --- & --- & 278 & 472 & 333 \\
ALB & 146 & 284 & 244 & 466 & 551 & 504 & 121 & 284 & 151 \\
ALP & 119 & 435 & 218 & 224 & 535 & 294 & 123 & 534 & 225 \\
ALT & 136 & 414 & 210 & 252 & 531 & 327 & 225 & 492 & 326 \\
AST & 71 & 383 & 142 & 186 & 441 & 240 & 45 & 396 & 118 \\
BUN & 235 & 458 & 267 & 360 & 546 & 400 & 254 & 479 & 285 \\
\bottomrule
\end{tabular}%
}
\label{tab:prevalence_interpret}
\end{table}
\clearpage

\begin{table}[ht]
\centering
\caption{\textbf{In-hospital mortality prediction performance in the eICU Collaborative Research Database.} Per-analyte precision, sensitivity, specificity, and balanced accuracy for Per$_{\mathrm{RI}}$ and NORMA$_{\mathrm{RI}}$ abnormality flags, restricted to measurements classified as normal by Pop$_{\mathrm{RI}}$. Analytes with fewer than 100 measurements are omitted (---).}
\adjustbox{max width=\textwidth}{%
\small
\begin{tabular}{lrrrrrrrrrr}
\toprule
 & &  & \multicolumn{4}{c}{Per$_{RI}$} & \multicolumn{4}{c}{NORMA$_{RI}$} \\
\cmidrule(lr){4-7} \cmidrule(lr){8-11}
Analyte & N & Events & Precision & Sensitivity & Specificity & Accuracy & Precision & Sensitivity & Specificity & Accuracy \\
\midrule
A1C & --- & --- & --- & --- & --- & --- & --- & --- & --- & --- \\
ALB & 891 & 49 & 0.05 & 0.20 & 0.75 & 0.48 & 0.08 & 0.18 & 0.88 & 0.53 \\
ALP & 8,843 & 1,143 & 0.13 & 0.60 & 0.42 & 0.51 & 0.16 & 0.21 & 0.83 & 0.52 \\
ALT & 7,391 & 913 & 0.12 & 0.56 & 0.43 & 0.50 & 0.12 & 0.20 & 0.80 & 0.50 \\
AST & 8,035 & 793 & 0.10 & 0.53 & 0.48 & 0.50 & 0.11 & 0.17 & 0.85 & 0.51 \\
BUN & 19,881 & 1,058 & 0.06 & 0.48 & 0.54 & 0.51 & 0.06 & 0.15 & 0.87 & 0.51 \\
\bottomrule
\end{tabular}%
}
\label{tab:eicu_eval_mortality}
\end{table}
\clearpage

\begin{table}[ht]
\centering
\caption{\textbf{Acute kidney injury prediction performance in the eICU Collaborative Research Database.} Per-analyte precision, sensitivity, specificity, and balanced accuracy for Per$_{\mathrm{RI}}$ and NORMA$_{\mathrm{RI}}$ abnormality flags, restricted to measurements classified as normal by Pop$_{\mathrm{RI}}$. Analytes with fewer than 100 measurements are omitted (---).}
\adjustbox{max width=\textwidth}{%
\small
\begin{tabular}{lrrrrrrrrrr}
\toprule
 & &  & \multicolumn{4}{c}{Per$_{RI}$} & \multicolumn{4}{c}{NORMA$_{RI}$} \\
\cmidrule(lr){4-7} \cmidrule(lr){8-11}
Analyte & N & Events & Precision & Sensitivity & Specificity & Accuracy & Precision & Sensitivity & Specificity & Accuracy \\
\midrule
A1C & --- & --- & --- & --- & --- & --- & --- & --- & --- & --- \\
ALB & 891 & 130 & 0.14 & 0.23 & 0.75 & 0.49 & 0.17 & 0.15 & 0.88 & 0.51 \\
ALP & 8,843 & 1,712 & 0.19 & 0.58 & 0.41 & 0.50 & 0.21 & 0.18 & 0.83 & 0.51 \\
ALT & 7,391 & 1,408 & 0.17 & 0.51 & 0.42 & 0.47 & 0.18 & 0.19 & 0.80 & 0.49 \\
AST & 8,035 & 1,384 & 0.17 & 0.50 & 0.47 & 0.49 & 0.16 & 0.14 & 0.85 & 0.49 \\
BUN & 19,881 & 1,164 & 0.05 & 0.38 & 0.53 & 0.46 & 0.05 & 0.12 & 0.87 & 0.49 \\
\bottomrule
\end{tabular}%
}
\label{tab:eicu_eval_aki}
\end{table}
\clearpage

\begin{table}[ht]
\centering
\caption{\textbf{Sepsis prediction performance in the eICU Collaborative Research Database.} Per-analyte precision, sensitivity, specificity, and balanced accuracy for Per$_{\mathrm{RI}}$ and NORMA$_{\mathrm{RI}}$ abnormality flags, restricted to measurements classified as normal by Pop$_{\mathrm{RI}}$. Analytes with fewer than 100 measurements are omitted (---).}
\adjustbox{max width=\textwidth}{%
\small
\begin{tabular}{lrrrrrrrrrr}
\toprule
 & &  & \multicolumn{4}{c}{Per$_{RI}$} & \multicolumn{4}{c}{NORMA$_{RI}$} \\
\cmidrule(lr){4-7} \cmidrule(lr){8-11}
Analyte & N & Events & Precision & Sensitivity & Specificity & Accuracy & Precision & Sensitivity & Specificity & Accuracy \\
\midrule
A1C & --- & --- & --- & --- & --- & --- & --- & --- & --- & --- \\
ALB & 891 & 87 & 0.08 & 0.20 & 0.75 & 0.47 & 0.13 & 0.16 & 0.88 & 0.52 \\
ALP & 8,843 & 2,106 & 0.23 & 0.57 & 0.41 & 0.49 & 0.23 & 0.16 & 0.83 & 0.50 \\
ALT & 7,391 & 1,925 & 0.25 & 0.54 & 0.42 & 0.48 & 0.25 & 0.19 & 0.80 & 0.49 \\
AST & 8,035 & 2,036 & 0.24 & 0.49 & 0.47 & 0.48 & 0.23 & 0.14 & 0.85 & 0.49 \\
BUN & 19,881 & 2,759 & 0.12 & 0.40 & 0.53 & 0.47 & 0.12 & 0.11 & 0.87 & 0.49 \\
\bottomrule
\end{tabular}%
}
\label{tab:eicu_eval_sepsis}
\end{table}
\clearpage

\begin{table}[ht]
\centering
\caption{\textbf{Prolonged ICU stay prediction performance in the eICU Collaborative Research Database.} Per-analyte precision, sensitivity, specificity, and balanced accuracy for Per$_{\mathrm{RI}}$ and NORMA$_{\mathrm{RI}}$ abnormality flags, restricted to measurements classified as normal by Pop$_{\mathrm{RI}}$. Analytes with fewer than 100 measurements are omitted (---).}
\adjustbox{max width=\textwidth}{%
\small
\begin{tabular}{lrrrrrrrrrr}
\toprule
 & &  & \multicolumn{4}{c}{Per$_{RI}$} & \multicolumn{4}{c}{NORMA$_{RI}$} \\
\cmidrule(lr){4-7} \cmidrule(lr){8-11}
Analyte & N & Events & Precision & Sensitivity & Specificity & Accuracy & Precision & Sensitivity & Specificity & Accuracy \\
\midrule
A1C & --- & --- & --- & --- & --- & --- & --- & --- & --- & --- \\
ALB & 891 & 158 & 0.17 & 0.24 & 0.75 & 0.50 & 0.23 & 0.16 & 0.88 & 0.52 \\
ALP & 8,843 & 2,882 & 0.32 & 0.57 & 0.41 & 0.49 & 0.35 & 0.19 & 0.84 & 0.51 \\
ALT & 7,391 & 2,412 & 0.32 & 0.55 & 0.42 & 0.48 & 0.33 & 0.20 & 0.80 & 0.50 \\
AST & 8,035 & 2,561 & 0.30 & 0.48 & 0.46 & 0.47 & 0.31 & 0.14 & 0.85 & 0.50 \\
BUN & 19,881 & 3,352 & 0.15 & 0.41 & 0.53 & 0.47 & 0.19 & 0.15 & 0.87 & 0.51 \\
\bottomrule
\end{tabular}%
}
\label{tab:eicu_eval_prolonged_los}
\end{table}
\clearpage

\begin{table}[ht]
\centering
\caption{\textbf{In-hospital mortality prediction performance in the INSPIRE cohort.} Per-analyte precision, sensitivity, specificity, and balanced accuracy for Per$_{\mathrm{RI}}$ and NORMA$_{\mathrm{RI}}$ abnormality flags, restricted to measurements classified as normal by Pop$_{\mathrm{RI}}$. Analytes with fewer than 100 measurements are omitted (---).}
\adjustbox{max width=\textwidth}{%
\small
\begin{tabular}{lrrrrrrrrrr}
\toprule
 & &  & \multicolumn{4}{c}{Per$_{RI}$} & \multicolumn{4}{c}{NORMA$_{RI}$} \\
\cmidrule(lr){4-7} \cmidrule(lr){8-11}
Analyte & N & Events & Precision & Sensitivity & Specificity & Accuracy & Precision & Sensitivity & Specificity & Accuracy \\
\midrule
A1C & 26 & 0 & 0.00 & --- & 0.73 & --- & 0.00 & --- & 0.92 & --- \\
ALB & 15,833 & 62 & 0.00 & 0.19 & 0.81 & 0.50 & 0.01 & 0.07 & 0.96 & 0.52 \\
ALP & 23,817 & 387 & 0.02 & 0.50 & 0.53 & 0.51 & 0.02 & 0.17 & 0.89 & 0.53 \\
ALT & 17,229 & 383 & 0.02 & 0.31 & 0.65 & 0.48 & 0.02 & 0.14 & 0.87 & 0.51 \\
AST & 24,951 & 513 & 0.03 & 0.45 & 0.63 & 0.54 & 0.03 & 0.12 & 0.92 & 0.52 \\
BUN & 18,100 & 225 & 0.01 & 0.32 & 0.69 & 0.51 & 0.03 & 0.10 & 0.96 & 0.53 \\
\bottomrule
\end{tabular}%
}
\label{tab:inspire_eval_mortality}
\end{table}
\clearpage

\begin{table}[ht]
\centering
\caption{\textbf{Perioperative infection prediction performance in the INSPIRE cohort.} Per-analyte precision, sensitivity, specificity, and balanced accuracy for Per$_{\mathrm{RI}}$ and NORMA$_{\mathrm{RI}}$ abnormality flags, restricted to measurements classified as normal by Pop$_{\mathrm{RI}}$. Analytes with fewer than 100 measurements are omitted (---).}
\adjustbox{max width=\textwidth}{%
\small
\begin{tabular}{lrrrrrrrrrr}
\toprule
 & &  & \multicolumn{4}{c}{Per$_{RI}$} & \multicolumn{4}{c}{NORMA$_{RI}$} \\
\cmidrule(lr){4-7} \cmidrule(lr){8-11}
Analyte & N & Events & Precision & Sensitivity & Specificity & Accuracy & Precision & Sensitivity & Specificity & Accuracy \\
\midrule
A1C & 26 & 2 & 0.14 & 0.50 & 0.75 & 0.62 & 0.00 & 0.00 & 0.92 & 0.46 \\
ALB & 15,833 & 437 & 0.03 & 0.17 & 0.81 & 0.49 & 0.03 & 0.04 & 0.96 & 0.50 \\
ALP & 23,817 & 1,174 & 0.04 & 0.37 & 0.52 & 0.44 & 0.04 & 0.10 & 0.89 & 0.49 \\
ALT & 17,229 & 1,005 & 0.05 & 0.29 & 0.64 & 0.46 & 0.05 & 0.12 & 0.87 & 0.49 \\
AST & 24,951 & 1,439 & 0.05 & 0.33 & 0.62 & 0.48 & 0.06 & 0.09 & 0.92 & 0.50 \\
BUN & 18,100 & 736 & 0.03 & 0.23 & 0.68 & 0.46 & 0.02 & 0.02 & 0.96 & 0.49 \\
\bottomrule
\end{tabular}%
}
\label{tab:inspire_eval_periop_infection}
\end{table}
\clearpage

\begin{table}[ht]
\centering
\caption{\textbf{Prolonged hospital stay prediction performance in the INSPIRE cohort.} Per-analyte precision, sensitivity, specificity, and balanced accuracy for Per$_{\mathrm{RI}}$ and NORMA$_{\mathrm{RI}}$ abnormality flags, restricted to measurements classified as normal by Pop$_{\mathrm{RI}}$. Analytes with fewer than 100 measurements are omitted (---).}
\adjustbox{max width=\textwidth}{%
\small
\begin{tabular}{lrrrrrrrrrr}
\toprule
 & &  & \multicolumn{4}{c}{Per$_{RI}$} & \multicolumn{4}{c}{NORMA$_{RI}$} \\
\cmidrule(lr){4-7} \cmidrule(lr){8-11}
Analyte & N & Events & Precision & Sensitivity & Specificity & Accuracy & Precision & Sensitivity & Specificity & Accuracy \\
\midrule
A1C & 26 & 19 & 0.71 & 0.26 & 0.71 & 0.49 & 1.00 & 0.10 & 1.00 & 0.55 \\
ALB & 15,833 & 10,599 & 0.68 & 0.19 & 0.81 & 0.50 & 0.66 & 0.03 & 0.96 & 0.50 \\
ALP & 23,817 & 18,548 & 0.78 & 0.48 & 0.53 & 0.50 & 0.70 & 0.10 & 0.85 & 0.47 \\
ALT & 17,229 & 13,384 & 0.76 & 0.34 & 0.61 & 0.48 & 0.74 & 0.13 & 0.84 & 0.48 \\
AST & 24,951 & 19,532 & 0.76 & 0.36 & 0.59 & 0.48 & 0.72 & 0.07 & 0.90 & 0.48 \\
BUN & 18,100 & 13,840 & 0.74 & 0.30 & 0.65 & 0.48 & 0.70 & 0.04 & 0.95 & 0.49 \\
\bottomrule
\end{tabular}%
}
\label{tab:inspire_eval_prolonged_los}
\end{table}
\clearpage

\begin{table}[ht]
\centering
\caption{\textbf{Unplanned ICU admission prediction performance in the INSPIRE cohort.} Per-analyte precision, sensitivity, specificity, and balanced accuracy for Per$_{\mathrm{RI}}$ and NORMA$_{\mathrm{RI}}$ abnormality flags, restricted to measurements classified as normal by Pop$_{\mathrm{RI}}$. Analytes with fewer than 100 measurements are omitted (---).}
\adjustbox{max width=\textwidth}{%
\small
\begin{tabular}{lrrrrrrrrrr}
\toprule
 & &  & \multicolumn{4}{c}{Per$_{RI}$} & \multicolumn{4}{c}{NORMA$_{RI}$} \\
\cmidrule(lr){4-7} \cmidrule(lr){8-11}
Analyte & N & Events & Precision & Sensitivity & Specificity & Accuracy & Precision & Sensitivity & Specificity & Accuracy \\
\midrule
A1C & 26 & 15 & 0.43 & 0.20 & 0.64 & 0.42 & 1.00 & 0.13 & 1.00 & 0.57 \\
ALB & 15,833 & 4,067 & 0.22 & 0.16 & 0.80 & 0.48 & 0.20 & 0.03 & 0.96 & 0.49 \\
ALP & 23,817 & 7,129 & 0.28 & 0.45 & 0.52 & 0.48 & 0.26 & 0.10 & 0.88 & 0.49 \\
ALT & 17,229 & 5,695 & 0.31 & 0.34 & 0.64 & 0.49 & 0.35 & 0.14 & 0.87 & 0.51 \\
AST & 24,951 & 7,856 & 0.27 & 0.32 & 0.60 & 0.46 & 0.25 & 0.06 & 0.92 & 0.49 \\
BUN & 18,100 & 5,436 & 0.29 & 0.30 & 0.68 & 0.49 & 0.33 & 0.04 & 0.96 & 0.50 \\
\bottomrule
\end{tabular}%
}
\label{tab:inspire_eval_unplanned_icu}
\end{table}
\clearpage

\begin{table}[ht]
\centering
\caption{\textbf{Positive predictive value across outcomes in Clalit Health Services.} Measurements classified as normal by Pop$_{\mathrm{RI}}$ are excluded. Per 100 patients flagged abnormal by Per$_{\mathrm{RI}}$ or NORMA$_{\mathrm{RI}}$, the number who experienced each outcome (anemia, chronic kidney disease, all-cause mortality, and type 2 diabetes), by analyte. Analytes with fewer than 100 measurements are omitted (---).}
\adjustbox{max width=\textwidth}{%
\small
\begin{tabular}{lrrrrrrrr}
\toprule
 & \multicolumn{2}{c}{Anemia} & \multicolumn{2}{c}{Chronic Kidney Disease} & \multicolumn{2}{c}{Mortality} & \multicolumn{2}{c}{Type 2 Diabetes} \\
\cmidrule(lr){2-3} \cmidrule(lr){4-5} \cmidrule(lr){6-7} \cmidrule(lr){8-9}
Analyte & Per$_{RI}$ & NORMA$_{RI}$ & Per$_{RI}$ & NORMA$_{RI}$ & Per$_{RI}$ & NORMA$_{RI}$ & Per$_{RI}$ & NORMA$_{RI}$ \\
\midrule
A1C & 21 & 30 & 39 & 48 & 23 & 34 & 70 & 78 \\
ALB & 21 & 25 & 44 & 51 & 26 & 33 & 78 & 81 \\
ALP & 19 & 24 & 40 & 47 & 25 & 30 & 74 & 79 \\
ALT & 19 & 22 & 40 & 43 & 24 & 27 & 74 & 78 \\
AST & 19 & 23 & 40 & 45 & 26 & 29 & 74 & 79 \\
BUN & 14 & 20 & 26 & 34 & 18 & 24 & 73 & 80 \\
CA & 21 & 27 & 45 & 53 & 28 & 35 & 77 & 82 \\
CL & 30 & 38 & 52 & 62 & 36 & 46 & 93 & 97 \\
CO2 & 100 & 100 & 75 & 100 & 50 & 80 & 100 & 100 \\
CRE & 16 & 24 & 34 & 57 & 21 & 32 & 75 & 84 \\
DBIL & --- & --- & --- & --- & --- & --- & --- & --- \\
GLU & 16 & 23 & 30 & 38 & 18 & 26 & 53 & 64 \\
HCT & 10 & 12 & 28 & 34 & 15 & 19 & 70 & 75 \\
HDL & 17 & 21 & 38 & 42 & 22 & 26 & 71 & 74 \\
HGB & 9 & 12 & 28 & 33 & 15 & 19 & 70 & 76 \\
K & 21 & 25 & 41 & 48 & 27 & 34 & 78 & 83 \\
LDL & 20 & 24 & 43 & 50 & 26 & 30 & 79 & 86 \\
MCH & 16 & 18 & 38 & 41 & 23 & 27 & 71 & 74 \\
MCHC & 14 & 17 & 30 & 34 & 18 & 22 & 73 & 77 \\
MCV & 17 & 20 & 37 & 40 & 23 & 26 & 73 & 75 \\
MPV & 19 & 18 & 37 & 36 & 24 & 24 & 74 & 73 \\
NA & 19 & 26 & 41 & 51 & 26 & 34 & 76 & 84 \\
PLT & 18 & 25 & 35 & 45 & 23 & 32 & 73 & 79 \\
RBC & 13 & 18 & 30 & 36 & 17 & 22 & 71 & 78 \\
RDW & 14 & 15 & 32 & 33 & 18 & 19 & 71 & 72 \\
TBIL & 24 & 22 & 46 & 44 & 30 & 29 & 79 & 77 \\
TC & 21 & 26 & 43 & 50 & 27 & 31 & 80 & 87 \\
TGL & --- & --- & --- & --- & --- & --- & --- & --- \\
TP & 23 & 27 & 47 & 54 & 30 & 37 & 78 & 81 \\
WBC & 18 & 22 & 37 & 41 & 23 & 27 & 73 & 78 \\
\midrule
\textbf{Overall} & \textbf{21} & \textbf{25} & \textbf{39} & \textbf{46} & \textbf{24} & \textbf{30} & \textbf{75} & \textbf{80} \\
\bottomrule
\end{tabular}%
}
\label{tab:chs_eval_interpret}
\end{table}
\clearpage

\begin{table}[ht]
\centering
\caption{\textbf{Positive predictive value across outcomes in the eICU Collaborative Research Database.} Measurements classified as normal by Pop$_{\mathrm{RI}}$ are excluded. Per 100 patients flagged abnormal by Per$_{\mathrm{RI}}$ or NORMA$_{\mathrm{RI}}$, the number who experienced each outcome (acute kidney injury, in-hospital mortality, prolonged ICU stay, and sepsis), by analyte. Analytes with fewer than 100 measurements are omitted (---).}
\adjustbox{max width=\textwidth}{%
\small
\begin{tabular}{lrrrrrrrr}
\toprule
 & \multicolumn{2}{c}{Acute Kidney Injury} & \multicolumn{2}{c}{Mortality} & \multicolumn{2}{c}{Prolonged LOS (>7d)} & \multicolumn{2}{c}{Sepsis} \\
\cmidrule(lr){2-3} \cmidrule(lr){4-5} \cmidrule(lr){6-7} \cmidrule(lr){8-9}
Analyte & Per$_{RI}$ & NORMA$_{RI}$ & Per$_{RI}$ & NORMA$_{RI}$ & Per$_{RI}$ & NORMA$_{RI}$ & Per$_{RI}$ & NORMA$_{RI}$ \\
\midrule
A1C & --- & --- & --- & --- & --- & --- & --- & --- \\
ALB & 14 & 17 & 5 & 8 & 17 & 23 & 8 & 13 \\
ALP & 19 & 21 & 13 & 16 & 32 & 35 & 23 & 23 \\
ALT & 17 & 18 & 12 & 12 & 32 & 33 & 25 & 25 \\
AST & 16 & 16 & 10 & 11 & 30 & 31 & 24 & 23 \\
BUN & 5 & 5 & 6 & 6 & 15 & 19 & 12 & 12 \\
CA & 13 & 15 & 8 & 8 & 20 & 23 & 13 & 14 \\
CL & 13 & 16 & 8 & 10 & 15 & 23 & 14 & 16 \\
CO2 & 9 & 11 & 6 & 8 & 12 & 23 & 11 & 12 \\
CRE & 7 & 5 & 8 & 10 & 19 & 19 & 14 & 13 \\
DBIL & 36 & 33 & 6 & 4 & 56 & 50 & 32 & 33 \\
GLU & 11 & 16 & 4 & 8 & 6 & 13 & 13 & 18 \\
HCT & 10 & 10 & 10 & 13 & 16 & 17 & 11 & 12 \\
HDL & --- & --- & --- & --- & --- & --- & --- & --- \\
HGB & 9 & 8 & 9 & 12 & 13 & 15 & 10 & 12 \\
K & 14 & 17 & 11 & 12 & 20 & 26 & 16 & 17 \\
LDL & --- & --- & --- & --- & --- & --- & --- & --- \\
MCH & 16 & 17 & 11 & 12 & 25 & 28 & 18 & 19 \\
MCHC & 14 & 14 & 9 & 10 & 18 & 19 & 15 & 15 \\
MCV & 15 & 16 & 11 & 14 & 22 & 27 & 18 & 19 \\
MPV & 15 & 15 & 12 & 11 & 24 & 24 & 16 & 16 \\
NA & 14 & 15 & 10 & 10 & 17 & 21 & 15 & 17 \\
PLT & 13 & 13 & 7 & 7 & 22 & 27 & 16 & 17 \\
RBC & 11 & 11 & 10 & 10 & 18 & 19 & 13 & 10 \\
RDW & 10 & 11 & 6 & 6 & 19 & 20 & 12 & 12 \\
TBIL & 19 & 20 & 14 & 15 & 35 & 35 & 26 & 25 \\
TC & 29 & 33 & 29 & 67 & 0 & 33 & 29 & 0 \\
TGL & 21 & 22 & 20 & 22 & 66 & 59 & 35 & 30 \\
TP & 19 & 22 & 11 & 11 & 33 & 37 & 24 & 26 \\
WBC & 12 & 14 & 7 & 8 & 17 & 20 & 12 & 12 \\
\midrule
\textbf{Overall} & \textbf{15} & \textbf{16} & \textbf{10} & \textbf{13} & \textbf{23} & \textbf{27} & \textbf{18} & \textbf{17} \\
\bottomrule
\end{tabular}%
}
\label{tab:eicu_eval_interpret}
\end{table}
\clearpage

\begin{table}[ht]
\centering
\caption{\textbf{Positive predictive value across outcomes in the INSPIRE cohort.} Measurements classified as normal by Pop$_{\mathrm{RI}}$ are excluded. Per 100 patients flagged abnormal by Per$_{\mathrm{RI}}$ or NORMA$_{\mathrm{RI}}$, the number who experienced each outcome (in-hospital mortality, perioperative infection, prolonged hospital stay, and unplanned ICU admission), by analyte. Analytes with fewer than 100 measurements are omitted (---).}
\adjustbox{max width=\textwidth}{%
\small
\begin{tabular}{lrrrrrrrr}
\toprule
 & \multicolumn{2}{c}{Mortality} & \multicolumn{2}{c}{Perioperative Infection} & \multicolumn{2}{c}{Prolonged LOS (>7d)} & \multicolumn{2}{c}{Unplanned ICU Admission} \\
\cmidrule(lr){2-3} \cmidrule(lr){4-5} \cmidrule(lr){6-7} \cmidrule(lr){8-9}
Analyte & Per$_{RI}$ & NORMA$_{RI}$ & Per$_{RI}$ & NORMA$_{RI}$ & Per$_{RI}$ & NORMA$_{RI}$ & Per$_{RI}$ & NORMA$_{RI}$ \\
\midrule
A1C & 0 & 0 & 14 & 0 & 71 & 100 & 43 & 100 \\
ALB & 0 & 1 & 2 & 3 & 68 & 66 & 22 & 20 \\
ALP & 2 & 2 & 4 & 4 & 78 & 70 & 28 & 26 \\
ALT & 2 & 2 & 5 & 5 & 76 & 74 & 31 & 35 \\
AST & 2 & 3 & 5 & 6 & 76 & 72 & 27 & 25 \\
BUN & 1 & 3 & 3 & 2 & 74 & 70 & 29 & 33 \\
CA & 2 & 7 & 3 & 6 & 67 & 65 & 23 & 20 \\
CL & 1 & 4 & 3 & 4 & 74 & 73 & 19 & 34 \\
CO2 & 4 & 8 & 9 & 12 & 83 & 89 & 51 & 57 \\
CRE & 3 & 7 & 4 & 5 & 70 & 67 & 21 & 19 \\
DBIL & --- & --- & --- & --- & --- & --- & --- & --- \\
GLU & 2 & 2 & 4 & 6 & 67 & 74 & 14 & 22 \\
HCT & 1 & 1 & 2 & 2 & 52 & 49 & 15 & 25 \\
HDL & --- & --- & --- & --- & --- & --- & --- & --- \\
HGB & 1 & 1 & 2 & 3 & 60 & 52 & 21 & 24 \\
K & 3 & 7 & 4 & 7 & 71 & 72 & 25 & 32 \\
LDL & --- & --- & --- & --- & --- & --- & --- & --- \\
MCH & --- & --- & --- & --- & --- & --- & --- & --- \\
MCHC & --- & --- & --- & --- & --- & --- & --- & --- \\
MCV & --- & --- & --- & --- & --- & --- & --- & --- \\
MPV & --- & --- & --- & --- & --- & --- & --- & --- \\
NA & 2 & 7 & 3 & 7 & 69 & 68 & 20 & 32 \\
PLT & 1 & 1 & 3 & 3 & 71 & 68 & 20 & 22 \\
RBC & --- & --- & --- & --- & --- & --- & --- & --- \\
RDW & --- & --- & --- & --- & --- & --- & --- & --- \\
TBIL & 3 & 2 & 7 & 5 & 75 & 70 & 28 & 27 \\
TC & --- & --- & --- & --- & --- & --- & --- & --- \\
TGL & --- & --- & --- & --- & --- & --- & --- & --- \\
TP & 1 & 1 & 4 & 8 & 73 & 74 & 22 & 22 \\
WBC & 1 & 2 & 3 & 4 & 69 & 76 & 19 & 23 \\
\midrule
\textbf{Overall} & \textbf{2} & \textbf{3} & \textbf{4} & \textbf{5} & \textbf{71} & \textbf{71} & \textbf{25} & \textbf{31} \\
\bottomrule
\end{tabular}%
}
\label{tab:inspire_eval_interpret}
\end{table}
\clearpage

\begin{table}[ht]
\centering
\caption{\textbf{Lead time for NORMA$_{\mathrm{RI}}$ abnormality detection relative to Pop$_{\mathrm{RI}}$.} Per analyte and cohort: number of tests flagged by NORMA$_{\mathrm{RI}}$ alone (NORMA$_{\mathrm{RI}}$-only) and the subset later confirmed by Pop$_{\mathrm{RI}}$ (later population), with median lead time and interquartile range (hours for eICU-CRD and INSPIRE, months for CHS).}
\adjustbox{max width=\textwidth}{%
\small
\begin{tabular}{lrrrrrrrrrrrr}
\toprule
 & \multicolumn{4}{c}{eICU-CRD} & \multicolumn{4}{c}{CHS} & \multicolumn{4}{c}{INSPIRE} \\
\cmidrule(lr){2-5}
\cmidrule(lr){6-9}
\cmidrule(lr){10-13}
Analyte & NORMA-only & Later Pop$_{RI}$ & Median (h) & IQR (h) & NORMA-only & Later Pop$_{RI}$ & Median (mo) & IQR (mo) & NORMA-only & Later Pop$_{RI}$ & Median (h) & IQR (h) \\
\midrule
A1C & --- & --- & --- & --- & 14,227 & 5,064 & 12.4 & 5.7--27.5 & 24 & 0 & --- & --- \\
ALB & 2,185 & 865 & 30.0 & 23.1--73.9 & 807,619 & 436,471 & 7.9 & 1.2--29.9 & 1,493 & 29 & 23.5 & 13.3--25.7 \\
ALP & 4,362 & 757 & 49.6 & 24.2--120.2 & 1,196,400 & 543,146 & 15.5 & 3.1--50.6 & 3,442 & 201 & 60.4 & 24.7--102.3 \\
ALT & 6,749 & 1,235 & 47.6 & 23.9--99.4 & 1,199,702 & 703,509 & 11.9 & 2.4--41.0 & 3,210 & 187 & 47.6 & 19.2--109.0 \\
AST & 7,720 & 1,317 & 47.4 & 23.9--95.5 & 1,195,620 & 551,419 & 10.0 & 1.6--38.0 & 2,204 & 29 & 16.6 & 8.2--24.0 \\
BUN & 25,578 & 7,039 & 28.0 & 23.1--71.3 & 568,660 & 402,995 & 7.6 & 1.1--29.2 & 1,983 & 409 & 24.9 & 13.0--55.2 \\
\bottomrule
\end{tabular}%
}
\label{tab:lead_time}
\end{table}
\clearpage

\begin{table}[ht]
\centering
\caption{\textbf{Proportional hazards analysis for all-cause mortality in Clalit Health Services.} Per-analyte hazard ratios (with 95\% confidence intervals) and concordance indices for Pop$_{\mathrm{RI}}$, Per$_{\mathrm{RI}}$, and NORMA$_{\mathrm{RI}}$ abnormality flags, adjusted for age and sex.}
\adjustbox{max width=\textwidth}{%
\small
\begin{tabular}{lrrrrrrrr}
\toprule
 & &  & \multicolumn{2}{c}{Pop$_{RI}$} & \multicolumn{2}{c}{Per$_{RI}$} & \multicolumn{2}{c}{NORMA$_{RI}$} \\
\cmidrule(lr){4-5} \cmidrule(lr){6-7} \cmidrule(lr){8-9}
Analyte & N train (events) & N test (events) & HR [95\% CI] & C & HR [95\% CI] & C & HR [95\% CI] & C \\
\midrule
A1C & 9517 (3118) & 6345 (2079) & 0.84 [0.65, 1.09] & 0.689 & 0.58 [0.54, 0.63] & 0.699 & 0.82 [0.65, 1.03] & 0.688 \\
ALB & 13612 (4221) & 9076 (2815) & 4.15 [3.83, 4.49] & 0.785 & 1.21 [1.10, 1.33] & 0.740 & 3.11 [2.80, 3.46] & 0.757 \\
ALP & 20262 (5615) & 13508 (3743) & 2.05 [1.94, 2.17] & 0.760 & 0.96 [0.87, 1.05] & 0.741 & 2.01 [1.86, 2.18] & 0.749 \\
ALT & 25254 (6667) & 16836 (4445) & 2.08 [1.95, 2.22] & 0.761 & 0.94 [0.86, 1.03] & 0.749 & 1.35 [1.27, 1.43] & 0.752 \\
AST & 25296 (6662) & 16864 (4442) & 1.85 [1.76, 1.94] & 0.758 & 1.02 [0.94, 1.12] & 0.744 & 1.55 [1.47, 1.65] & 0.750 \\
BUN & 26980 (7187) & 17987 (4792) & 2.82 [2.53, 3.15] & 0.747 & 0.81 [0.75, 0.88] & 0.739 & 2.58 [2.31, 2.87] & 0.746 \\
\bottomrule
\end{tabular}%
}
\label{tab:chs_cox_mortality}
\end{table}
\clearpage

\begin{table}[ht]
\centering
\caption{\textbf{Proportional hazards analysis for in-hospital mortality in the eICU Collaborative Research Database.} Per-analyte hazard ratios (with 95\% confidence intervals) and concordance indices for Pop$_{\mathrm{RI}}$, Per$_{\mathrm{RI}}$, and NORMA$_{\mathrm{RI}}$ abnormality flags, adjusted for age and sex.}
\adjustbox{max width=\textwidth}{%
\small
\begin{tabular}{lrrrrrrrr}
\toprule
 & &  & \multicolumn{2}{c}{Pop$_{RI}$} & \multicolumn{2}{c}{Per$_{RI}$} & \multicolumn{2}{c}{NORMA$_{RI}$} \\
\cmidrule(lr){4-5} \cmidrule(lr){6-7} \cmidrule(lr){8-9}
Analyte & N train (events) & N test (events) & HR [95\% CI] & C & HR [95\% CI] & C & HR [95\% CI] & C \\
\midrule
A1C & --- & --- & --- & --- & --- & --- & --- & --- \\
ALB & 11151 (1560) & 7435 (1040) & 1.04 [0.75, 1.43] & 0.544 & 1.01 [0.65, 1.57] & 0.544 & 1.14 [0.78, 1.65] & 0.544 \\
ALP & 7966 (1118) & 5311 (746) & 0.97 [0.86, 1.09] & 0.534 & 1.09 [0.93, 1.28] & 0.534 & 1.12 [0.99, 1.26] & 0.535 \\
ALT & 7288 (976) & 4859 (650) & 0.98 [0.86, 1.11] & 0.538 & 1.05 [0.89, 1.24] & 0.539 & 0.99 [0.87, 1.13] & 0.537 \\
AST & 7446 (912) & 4964 (608) & 1.68 [1.47, 1.93] & 0.571 & 1.82 [1.52, 2.19] & 0.562 & 1.61 [1.40, 1.86] & 0.575 \\
BUN & 33379 (3378) & 22253 (2252) & 1.74 [1.58, 1.92] & 0.573 & 1.84 [1.59, 2.14] & 0.559 & 1.77 [1.59, 1.97] & 0.571 \\
\bottomrule
\end{tabular}%
}
\label{tab:eicu_cox_mortality}
\end{table}
\clearpage

\begin{table}[ht]
\centering
\caption{\textbf{Proportional hazards analysis for acute kidney injury in the eICU Collaborative Research Database.} Per-analyte hazard ratios (with 95\% confidence intervals) and concordance indices for Pop$_{\mathrm{RI}}$, Per$_{\mathrm{RI}}$, and NORMA$_{\mathrm{RI}}$ abnormality flags, adjusted for age and sex.}
\adjustbox{max width=\textwidth}{%
\small
\begin{tabular}{lrrrrrrrr}
\toprule
 & &  & \multicolumn{2}{c}{Pop$_{RI}$} & \multicolumn{2}{c}{Per$_{RI}$} & \multicolumn{2}{c}{NORMA$_{RI}$} \\
\cmidrule(lr){4-5} \cmidrule(lr){6-7} \cmidrule(lr){8-9}
Analyte & N train (events) & N test (events) & HR [95\% CI] & C & HR [95\% CI] & C & HR [95\% CI] & C \\
\midrule
A1C & --- & --- & --- & --- & --- & --- & --- & --- \\
ALB & 11121 (2490) & 7414 (1660) & 1.28 [1.01, 1.63] & 0.516 & 1.45 [1.03, 2.03] & 0.515 & 1.36 [1.04, 1.79] & 0.515 \\
ALP & 7949 (1702) & 5300 (1134) & 1.25 [1.14, 1.37] & 0.544 & 1.12 [0.99, 1.26] & 0.516 & 1.24 [1.12, 1.36] & 0.534 \\
ALT & 7267 (1472) & 4846 (982) & 1.08 [0.97, 1.19] & 0.504 & 0.94 [0.82, 1.07] & 0.505 & 1.06 [0.95, 1.18] & 0.504 \\
AST & 7429 (1464) & 4954 (976) & 1.19 [1.08, 1.32] & 0.529 & 1.07 [0.95, 1.21] & 0.504 & 1.14 [1.03, 1.27] & 0.522 \\
BUN & 33279 (4964) & 22187 (3310) & 1.89 [1.76, 2.04] & 0.566 & 1.63 [1.47, 1.82] & 0.535 & 1.73 [1.60, 1.87] & 0.555 \\
\bottomrule
\end{tabular}%
}
\label{tab:eicu_cox_aki}
\end{table}
\clearpage

\begin{table}[ht]
\centering
\caption{\textbf{Proportional hazards analysis for sepsis in the eICU Collaborative Research Database.} Per-analyte hazard ratios (with 95\% confidence intervals) and concordance indices for Pop$_{\mathrm{RI}}$, Per$_{\mathrm{RI}}$, and NORMA$_{\mathrm{RI}}$ abnormality flags, adjusted for age and sex.}
\adjustbox{max width=\textwidth}{%
\small
\begin{tabular}{lrrrrrrrr}
\toprule
 & &  & \multicolumn{2}{c}{Pop$_{RI}$} & \multicolumn{2}{c}{Per$_{RI}$} & \multicolumn{2}{c}{NORMA$_{RI}$} \\
\cmidrule(lr){4-5} \cmidrule(lr){6-7} \cmidrule(lr){8-9}
Analyte & N train (events) & N test (events) & HR [95\% CI] & C & HR [95\% CI] & C & HR [95\% CI] & C \\
\midrule
A1C & --- & --- & --- & --- & --- & --- & --- & --- \\
ALB & 11106 (2754) & 7405 (1836) & 2.27 [1.72, 3.00] & 0.519 & 2.42 [1.64, 3.56] & 0.518 & 2.24 [1.66, 3.02] & 0.518 \\
ALP & 7938 (2020) & 5292 (1347) & 1.17 [1.07, 1.28] & 0.544 & 0.98 [0.88, 1.10] & 0.503 & 1.12 [1.02, 1.22] & 0.532 \\
ALT & 7256 (1877) & 4838 (1252) & 1.01 [0.92, 1.11] & 0.510 & 1.01 [0.90, 1.13] & 0.510 & 0.99 [0.90, 1.09] & 0.512 \\
AST & 7415 (1904) & 4944 (1270) & 1.08 [0.98, 1.18] & 0.502 & 0.96 [0.87, 1.07] & 0.517 & 1.08 [0.98, 1.18] & 0.495 \\
BUN & 33240 (5690) & 22161 (3793) & 1.22 [1.15, 1.30] & 0.533 & 1.08 [0.99, 1.17] & 0.527 & 1.16 [1.09, 1.24] & 0.527 \\
\bottomrule
\end{tabular}%
}
\label{tab:eicu_cox_sepsis}
\end{table}
\clearpage

\begin{table}[ht]
\centering
\caption{\textbf{Proportional hazards analysis for prolonged ICU stay in the eICU Collaborative Research Database.} Prolonged ICU stay is defined as ICU stay greater than 7 days. Per-analyte hazard ratios (with 95\% confidence intervals) and concordance indices for Pop$_{\mathrm{RI}}$, Per$_{\mathrm{RI}}$, and NORMA$_{\mathrm{RI}}$ abnormality flags, adjusted for age and sex.}
\adjustbox{max width=\textwidth}{%
\small
\begin{tabular}{lrrrrrrrr}
\toprule
 & &  & \multicolumn{2}{c}{Pop$_{RI}$} & \multicolumn{2}{c}{Per$_{RI}$} & \multicolumn{2}{c}{NORMA$_{RI}$} \\
\cmidrule(lr){4-5} \cmidrule(lr){6-7} \cmidrule(lr){8-9}
Analyte & N train (events) & N test (events) & HR [95\% CI] & C & HR [95\% CI] & C & HR [95\% CI] & C \\
\midrule
A1C & --- & --- & --- & --- & --- & --- & --- & --- \\
ALB & 11140 (3587) & 7428 (2392) & 0.75 [0.61, 0.93] & 0.535 & 0.70 [0.51, 0.96] & 0.536 & 0.77 [0.60, 0.97] & 0.535 \\
ALP & 7956 (2821) & 5305 (1881) & 0.68 [0.63, 0.73] & 0.551 & 0.84 [0.76, 0.92] & 0.533 & 0.73 [0.68, 0.79] & 0.548 \\
ALT & 7276 (2596) & 4851 (1731) & 0.82 [0.76, 0.88] & 0.533 & 0.87 [0.78, 0.96] & 0.520 & 0.85 [0.78, 0.92] & 0.527 \\
AST & 7437 (2658) & 4958 (1772) & 0.86 [0.80, 0.93] & 0.545 & 0.98 [0.89, 1.07] & 0.534 & 0.87 [0.81, 0.94] & 0.545 \\
BUN & 33336 (7040) & 22225 (4694) & 0.66 [0.62, 0.71] & 0.550 & 0.76 [0.70, 0.83] & 0.535 & 0.67 [0.63, 0.72] & 0.548 \\
\bottomrule
\end{tabular}%
}
\label{tab:eicu_cox_prolonged_los}
\end{table}
\clearpage

\begin{table}[ht]
\centering
\caption{\textbf{Proportional hazards analysis for in-hospital mortality in the INSPIRE cohort.} Per-analyte hazard ratios (with 95\% confidence intervals) and concordance indices for Pop$_{\mathrm{RI}}$, Per$_{\mathrm{RI}}$, and NORMA$_{\mathrm{RI}}$ abnormality flags, adjusted for age and sex.}
\adjustbox{max width=\textwidth}{%
\small
\begin{tabular}{lrrrrrrrr}
\toprule
 & &  & \multicolumn{2}{c}{Pop$_{RI}$} & \multicolumn{2}{c}{Per$_{RI}$} & \multicolumn{2}{c}{NORMA$_{RI}$} \\
\cmidrule(lr){4-5} \cmidrule(lr){6-7} \cmidrule(lr){8-9}
Analyte & N train (events) & N test (events) & HR [95\% CI] & C & HR [95\% CI] & C & HR [95\% CI] & C \\
\midrule
A1C & 233 (13) & 156 (8) & 0.13 [0.01, 2.06] & 0.446 & 0.13 [0.01, 2.06] & 0.446 & --- & 0.482 \\
ALB & 18427 (470) & 12285 (314) & 6.58 [4.90, 8.83] & 0.805 & 4.28 [3.06, 6.00] & 0.719 & 5.87 [4.30, 8.02] & 0.785 \\
ALP & 16668 (437) & 11113 (292) & 2.10 [1.72, 2.56] & 0.745 & 2.40 [1.89, 3.04] & 0.727 & 2.43 [1.98, 2.98] & 0.754 \\
ALT & 15676 (441) & 10452 (294) & 1.66 [1.37, 2.02] & 0.632 & 1.48 [1.20, 1.82] & 0.636 & 1.51 [1.24, 1.83] & 0.637 \\
AST & 15814 (415) & 10544 (277) & 2.17 [1.76, 2.68] & 0.742 & 1.99 [1.62, 2.45] & 0.713 & 2.21 [1.81, 2.70] & 0.732 \\
BUN & 17938 (447) & 11960 (298) & 2.40 [1.90, 3.02] & 0.724 & 2.21 [1.71, 2.86] & 0.670 & 2.41 [1.91, 3.06] & 0.707 \\
\bottomrule
\end{tabular}%
}
\label{tab:inspire_cox_mortality}
\end{table}
\clearpage

\begin{table}[ht]
\centering
\caption{\textbf{Proportional hazards analysis for perioperative infection in the INSPIRE cohort.} Per-analyte hazard ratios (with 95\% confidence intervals) and concordance indices for Pop$_{\mathrm{RI}}$, Per$_{\mathrm{RI}}$, and NORMA$_{\mathrm{RI}}$ abnormality flags, adjusted for age and sex.}
\adjustbox{max width=\textwidth}{%
\small
\begin{tabular}{lrrrrrrrr}
\toprule
 & &  & \multicolumn{2}{c}{Pop$_{RI}$} & \multicolumn{2}{c}{Per$_{RI}$} & \multicolumn{2}{c}{NORMA$_{RI}$} \\
\cmidrule(lr){4-5} \cmidrule(lr){6-7} \cmidrule(lr){8-9}
Analyte & N train (events) & N test (events) & HR [95\% CI] & C & HR [95\% CI] & C & HR [95\% CI] & C \\
\midrule
A1C & 226 (32) & 152 (21) & 0.82 [0.35, 1.97] & 0.559 & 1.30 [0.43, 4.00] & 0.572 & 1.80 [0.58, 5.56] & 0.572 \\
ALB & 18112 (691) & 12076 (461) & 1.22 [1.05, 1.43] & 0.624 & 1.14 [0.97, 1.34] & 0.626 & 1.15 [0.98, 1.34] & 0.631 \\
ALP & 16373 (656) & 10916 (438) & 1.16 [0.98, 1.38] & 0.593 & 1.12 [0.96, 1.30] & 0.598 & 1.16 [0.99, 1.36] & 0.590 \\
ALT & 15382 (665) & 10255 (443) & 1.03 [0.88, 1.21] & 0.610 & 1.02 [0.87, 1.19] & 0.610 & 1.08 [0.93, 1.26] & 0.614 \\
AST & 15519 (665) & 10347 (444) & 1.45 [1.15, 1.84] & 0.642 & 1.20 [1.02, 1.40] & 0.625 & 1.26 [1.03, 1.54] & 0.644 \\
BUN & 17637 (682) & 11758 (455) & 1.09 [0.94, 1.27] & 0.557 & 1.06 [0.90, 1.24] & 0.548 & 1.12 [0.96, 1.31] & 0.556 \\
\bottomrule
\end{tabular}%
}
\label{tab:inspire_cox_periop_infection}
\end{table}
\clearpage

\begin{table}[ht]
\centering
\caption{\textbf{Proportional hazards analysis for prolonged hospital stay in the INSPIRE cohort.} Prolonged hospital stay is defined as hospital stay greater than 7 days. Per-analyte hazard ratios (with 95\% confidence intervals) and concordance indices for Pop$_{\mathrm{RI}}$, Per$_{\mathrm{RI}}$, and NORMA$_{\mathrm{RI}}$ abnormality flags, adjusted for age and sex.}
\adjustbox{max width=\textwidth}{%
\small
\begin{tabular}{lrrrrrrrr}
\toprule
 & &  & \multicolumn{2}{c}{Pop$_{RI}$} & \multicolumn{2}{c}{Per$_{RI}$} & \multicolumn{2}{c}{NORMA$_{RI}$} \\
\cmidrule(lr){4-5} \cmidrule(lr){6-7} \cmidrule(lr){8-9}
Analyte & N train (events) & N test (events) & HR [95\% CI] & C & HR [95\% CI] & C & HR [95\% CI] & C \\
\midrule
A1C & 233 (164) & 156 (109) & 0.95 [0.62, 1.45] & 0.511 & 0.82 [0.46, 1.45] & 0.500 & 0.88 [0.52, 1.48] & 0.523 \\
ALB & 18427 (14140) & 12285 (9427) & 0.70 [0.67, 0.73] & 0.566 & 0.73 [0.71, 0.76] & 0.568 & 0.66 [0.64, 0.69] & 0.572 \\
ALP & 16668 (13163) & 11113 (8776) & 0.67 [0.64, 0.71] & 0.558 & 1.10 [1.06, 1.13] & 0.532 & 0.82 [0.79, 0.85] & 0.538 \\
ALT & 15676 (12343) & 10452 (8230) & 0.85 [0.82, 0.88] & 0.545 & 1.02 [0.99, 1.06] & 0.533 & 0.92 [0.89, 0.95] & 0.536 \\
AST & 15814 (12427) & 10544 (8286) & 0.71 [0.65, 0.77] & 0.539 & 1.12 [1.08, 1.16] & 0.539 & 0.95 [0.90, 1.00] & 0.530 \\
BUN & 17938 (14031) & 11960 (9355) & 0.75 [0.72, 0.77] & 0.553 & 0.96 [0.92, 0.99] & 0.538 & 0.76 [0.74, 0.79] & 0.551 \\
\bottomrule
\end{tabular}%
}
\label{tab:inspire_cox_prolonged_los}
\end{table}
\clearpage

\begin{table}[ht]
\centering
\caption{\textbf{Proportional hazards analysis for unplanned ICU admission in the INSPIRE cohort.} Per-analyte hazard ratios (with 95\% confidence intervals) and concordance indices for Pop$_{\mathrm{RI}}$, Per$_{\mathrm{RI}}$, and NORMA$_{\mathrm{RI}}$ abnormality flags, adjusted for age and sex.}
\adjustbox{max width=\textwidth}{%
\small
\begin{tabular}{lrrrrrrrr}
\toprule
 & &  & \multicolumn{2}{c}{Pop$_{RI}$} & \multicolumn{2}{c}{Per$_{RI}$} & \multicolumn{2}{c}{NORMA$_{RI}$} \\
\cmidrule(lr){4-5} \cmidrule(lr){6-7} \cmidrule(lr){8-9}
Analyte & N train (events) & N test (events) & HR [95\% CI] & C & HR [95\% CI] & C & HR [95\% CI] & C \\
\midrule
A1C & 233 (105) & 156 (70) & 1.09 [0.68, 1.74] & 0.495 & 1.00 [0.59, 1.69] & 0.499 & 1.17 [0.66, 2.07] & 0.528 \\
ALB & 18427 (5221) & 12285 (3481) & 0.85 [0.80, 0.91] & 0.541 & 0.90 [0.86, 0.96] & 0.542 & 0.94 [0.89, 1.00] & 0.539 \\
ALP & 16668 (5096) & 11113 (3397) & 1.07 [0.99, 1.15] & 0.535 & 0.96 [0.91, 1.01] & 0.539 & 1.01 [0.95, 1.08] & 0.536 \\
ALT & 15676 (5066) & 10452 (3378) & 0.95 [0.90, 1.02] & 0.535 & 0.94 [0.88, 0.99] & 0.537 & 1.01 [0.96, 1.07] & 0.534 \\
AST & 15814 (5026) & 10544 (3351) & 1.14 [1.01, 1.28] & 0.530 & 0.84 [0.79, 0.89] & 0.540 & 0.90 [0.82, 0.98] & 0.532 \\
BUN & 17938 (5376) & 11960 (3585) & 1.00 [0.94, 1.06] & 0.523 & 0.94 [0.89, 0.99] & 0.524 & 1.01 [0.96, 1.07] & 0.523 \\
\bottomrule
\end{tabular}%
}
\label{tab:inspire_cox_unplanned_icu}
\end{table}
\clearpage

\end{document}